\documentclass[article]{jss}

\usepackage{thumbpdf,lmodern}

\usepackage{framed}

\usepackage{amsmath, bm}
\usepackage{enumerate}
\usepackage{xcolor}

\author{Yang Qian \\Hefei University of Technology \And Yuanchun Jiang\thanks{Corresponding author.  E-mail address: ycjiang@hfut.edu.cn} \\Hefei University of Technology
    \AND Yidong Chai\\Hefei University of Technology
   \And Yezheng Liu\\Hefei University of Technology  \AND Jianshan Sun\\Hefei University of Technology}

\Plainauthor{Yuanchun Jiang, Second Author}

\title{TopicModel4J: A \proglang{Java} Package for Topic Models}
\Plaintitle{TopicModel4J: A Java Package for Topic Models}
\Shorttitle{TopicModel4J: A \proglang{Java} Package for Topic Models}

\Abstract{
  Topic models provide a flexible and principled framework for exploring hidden structure in high-dimensional co-occurrence data and are commonly used natural language processing (NLP)
  of text. In this paper, we design and implement a \proglang{Java} package, \pkg{TopicModel4J}, which contains 13 kinds of representative algorithms for fitting topic models.
  The \pkg{TopicModel4J} in the \proglang{Java} programming environment provides an easy-to-use interface for data analysts to run the algorithms, and allow to easily input and output data. In addition, this package provides a few unstructured text preprocessing techniques, such as splitting textual data into words,
   lowercasing the words, preforming lemmatization and removing the useless characters, URLs and stop words.
}

\Plainkeywords{natural language processing, topic model, Gibbs sampling, variational inference, Java}

\Address{
  Yang Qian \\
  School of Management \\
  Hefei University of Technology\\
  Hefei, Anhui 230009, China \\
  E-mail: \email{yqian@mail.hfut.edu.cn}\\
  \\
  Yuanchun Jiang \textit{(corresponding author)} \\
  School of Management \\
  Hefei University of Technology\\
  Hefei, Anhui 230009, China \\
  E-mail: \email{ycjiang@hfut.edu.cn}\\
}

\begin{document}


\section{Introduction} \label{sec:intro}

Topic models are a type of probabilistic generative models which are widely used to discover latent semantics (e.g., topics) from a large collection of documents. Based upon the intuition that one document can be represented as an admixture of abstract topics and one topic is a set of similar vocabulary words, topic models can find word patterns, abstract topics and topics that best characterize each document in an unsupervised manner. Therefore, topic models can provide insights for us to understand collections of documents, or any other data types as long as they can be treated as documents (e.g., images, networks, and genetic information).

Proposed by \cite{blei2003latent}, latent Dirichlet allocation (LDA) is the {most common and representative} topic model. The core idea of LDA is to introduce Dirichlet prior for document-topic and topic-word distributions (two multinomial distributions), thus to discover better latent semantics. Since then, numerious classical topic models, which are generally extensions on LDA, have been devised to better solve tasks from various domains. To encode authorship information, an important aspect for documents, \cite{rosen2004author} proposed the author-topic model where each author is associated with a distribution over topics. To make the training scale to large corpus, \cite{teh2007collapsed} proposed the collapsed variational Bayesian inference (CVB) algorithm to estimate model parameters. For short documents, which are highly prevalent in social medial, \cite{yin2014dirichlet} presented a collapsed Gibbs sampling algorithm for the Dirichlet Multinomial Mixture (DMM) model to cluster the short text. To address the limitation that the number of topics is predefined subjectively in LDA, \cite{teh2005sharing} proposed a nonparametric Bayesian method, Hierarchical Dirichlet Processes (HDP), where a hierarchy of Dirichlet processes were introduced to capture the uncertainty in the number of topics. Then, they used Gibbs sampler based on the Chinese restaurant franchise to learn model parameters. Such classfical topic models have been widely used in various research fields such as personalized recommendation \citep{ansari2018probabilistic,ling2014ratings,yuan2014generative}, information retrieval \citep{wang2015hierarchical,gerlach2018network}, marketing-operations management \citep{hannigan2019topic} and bio-medicine \citep{gonzalez2019cistopic}.

However, the implementation of topic models involves much programming work, which is laborious and time-consuming. To make it more  efficient and convenient, a range of packages have been developed for topic models. \pkg{GibbsLDA++} \citep{phan2007gibbslda++} was released under the GNU General Public License (GPL) and the \proglang{C++} was adopted to implement the LDA with Gibbs sampling (\url{http://gibbslda.sourceforge.net/}). David M. Blei released the \proglang{C} implementation of variational expectation-maximization (EM) for LDA under the GPL (\url{https://github.com/blei-lab/lda-c}). \cite{hornik2011topicmodels} provided an \proglang{R} package \pkg{topicmodels} to fit topic models with both variational EM algorithm and Gibbs sampling (\url{https://cran.r-project.org/web/packages/topicmodels}). To provides an interface to the code for fitting Hierarchical Dirichlet Processes \cite{teh2005sharing}, David M. Blei provided an implementation using \proglang{C++} scripts (\url{https://github.com/blei-lab/hdp}).

In this paper, we design and implement a \proglang{Java}-based package \pkg{TopicModel4J} for fitting a series of popular classical topic models. \pkg{TopicModel4J} provides an easy-to-use interface for parameter estimation and inference, which make it straightforward and simple for researchers to get started and to design their own models. In particular, for researchers and data analysts, they can directly call the algorithm using a few lines of codes. In addition, the required input for all algorithms in our package is a plain text file, freeing researchers from the input related work such as constructing vocabulary files, counting the number of words in each document, and creating document-term-count files. Here, we give a brief description of each algorithm in the package \pkg{TopicModel4J}.

\begin{itemize}
    \item LDA with the collapsed Gibbs sampling: Proposed by \cite{griffiths2004finding}, this algorithm used the collapsed Gibbs sampling to estimate the parameters for LDA.
    \item LDA with the collapsed variational Bayesian (CVB) inference: Considering the convergence and the computational efficiency of Gibbs sampling, Proposed by \cite{teh2007collapsed}, this algorithm combined the key insights of Gibbs sampling and standard variational inference algorithm, and then propose the CVB algorithm for inferring LDA.
    \item Sentence-LDA: Sentence-LDA is proposed by \cite{jo2011aspect}. Unlike LDA, Sentence-LDA assumes that the words in one sentences of the document is drawn from the same topic. This algorithm is often used to analyze the unstructured online reviews \citep{buschken2016sentence}. In the package \pkg{TopicModel4J}, we use collapsed Gibbs sampling for inferring the Sentence-LDA.
    \item HDP with Chinese restaurant franchise: HDP \citep{teh2005sharing} is nonparametric approach which extends the Dirichlet processes (DP) to model the uncertainty concerning the number of topics. HDP can be inferred by Gibbs sampling based on Chinese restaurant franchise metaphor.
    \item Dirichlet Multinomial Mixture (DMM) Model: DMM model \citep{nigam2000text} assumes that each document is only assigned to one cluster or topic. Recently, this model is widely used for short text clustering \citep{li2016topic,liang2016dynamic,yin2014dirichlet}. In the package \pkg{TopicModel4J}, we use collapsed Gibbs sampling algorithm for inferring the DMM model \citep{yin2014dirichlet}.
     \item Dirichlet Process Multinomial Mixture (DPMM) model: DPMM model \citep{yu2010document,zhang2005probabilistic} is an extension of DMM model. This model is often used to handle the short text clustering problem. Compared with DMM, the character of the DPMM model is that the number of clusters can be determined automatically. In our package, we use collapsed Gibbs sampling algorithm for inferring the parameters of DPMM \citep{yin2016model}.
     \item Pseudo-document-based Topic Model (PTM): PTM \citep{zuo2016topic} is a short text topic modeling, which leads into the thought of pseudo document to indirectly aggregate short texts against the sparsity in terms of word co-occurrences. In our package, we use collapsed Gibbs sampling algorithm for approximate inference of PTM.
     \item Biterm topic model (BTM): BTM \citep{cheng2014btm,yan2013biterm} is a popular short text topic modeling, which introduce the word pair co-occurring in a short document to overcome the sparsity of word co-occurrence. The word pair that is generated by the short document, is noted as 'biterm'. In our package, we conduct an approximate inference for BTM using the collapsed Gibbs sampling algorithm.
      \item Author-topic model (ATM): ATM \citep{rosen2004author} is an extension of LDA, that models the interests of authors and the documents at the same time. In our package, we use the collapsed Gibbs sampling algorithm to estimate the parameters of ATM.
      \item Link LDA: Link LDA \citep{erosheva2004mixed} is originally proposed to model the documents containing abstracts and references. The references of each document can be viewed as a set of links. Recently, different variants of Link LDA have been recently proposed to mining the topic-specific influencers [26-28]. In our package, we use collapsed Gibbs sampling algorithm to estimate the parameters of Link LDA \citep{su2018identifying,bi2014scalable,yang2009combining}.
      \item Labeled LDA: Labeled LDA is a supervised topic model that proposed by \cite{ramage2009labeled}. This model is applicable for multi-labeled corpora. Unlike LDA, Labeled LDA defines a consistent one-to-one match between the LDA's latent topics and the labels. Hence, this model enhances the results that can be interpretative. In our package, we use collapsed Gibbs sampling algorithm to estimate the parameters of Labeled LDA.
      \item Partially Labeled Dirichlet Allocation (PLDA): PLDA is also a supervised topic model that is proposed by \cite{ramage2011partially}. This model is applicable for multi-labeled corpora. Unlike Labeled LDA, PLDA introduce one class (containing multiple topics) for each label in the label set. In our package, we use collapsed Gibbs sampling algorithm to estimate the parameters of PLDA.
     \item Dual-Sparse Topic Model: This model is a sparsity-enhanced topic model, proposed by \cite{lin2014dual}. Unlike LDA, this model assumes that each document often concentrates on several salient topics and each topic often focus on a narrow range of words. In our package, we use CVB inference to estimate the parameters of the dual-sparse topic model.
\end{itemize}
The remainder of this paper is structured as follows: Section 2 reviews the related methods in \pkg{TopicModel4J}. In Section 3, we introduce the formula interface of each model in \pkg{TopicModel4J} and give the applications on real dataset. We finally present conclusions in Section 4.



\section{Models} \label{sec:models}

\subsection{LDA}
LDA \citep{blei2003latent} is a generative statistical model that depicts how a corpus of documents is generated by a set of latent topics. Let $M$ denote the number of documents in a collection; $V$ denotes the number of words in the vocabulary; $K$ denotes the number of topics. Further, we define $w_{m,n}$ as the $n$-th word (token) in the $m$-th document, $z_{m.n}\in \left \{1,,2,\cdots ,K  \right \}$ as the topic assignment for $w_{m,n}$, and $N_{m}$ as number of word tokens in document $m$. Let $\theta_{m,k}$ denote the probability of topic $k$ occurring in the $m$-th document and $\phi_{k,v}$ denote the probability of word $v$ occurring in topic $k$, then the model parameters $\bm{\theta _{m}}= \left \{ \theta _{m,k} \right \}_{k=1}^{K}, \forall m$ and $\bm{\phi _{k}}=\left \{ \phi _{k,v} \right \}_{v=1}^{V}, \forall k$ are defined as $K$-dimensional topic mixing vector and $V$-dimensional topic-word vector respectively. The generative process of LDA is described as follows:
\begin{enumerate}[1)]
    \item For each topic $k\in \left [ 1,K \right ]$
    \begin{enumerate}
        \item Draw $\bm{\phi _{k}}\sim Dirichlet\left ( \beta  \right )$
    \end{enumerate}
    \item For each document $m\in \left [ 1,M \right ]$
     \begin{enumerate}
        \item Draw $\bm{\theta _{m}}\sim Dirichlet\left ( \alpha  \right )$
        \item For each word $w_{m,n}$ in document $m$, $n\in \left [ 1,N_{m} \right ]$
        \begin{enumerate}
            \item Draw a topic $z_{m,n}\sim Multinomial\left ( \bm{\theta _{m}}  \right )$
            \item Draw a word $w_{m,n}\sim Multinomial\left ( \bm{\phi _{z_{m,n}}}  \right )$
        \end{enumerate}
    \end{enumerate}
\end{enumerate}
Where $Dirichlet\left ( \cdot \right )$ and $Multinomial\left ( \cdot \right )$ denote the Dirichlet distribution and Multinomial distribution, respectively. $\alpha$ and $\beta$ are hyperparameters for Dirichlet priors.
\subsubsection{Collapsed Gibbs Sampling for LDA}
As an extention and improvment of Gibbs sampling, the collapsed Gibbs sampling integrates out some variables from the joint distribution, enabling a more reliable parameter learning  by reducing estimation variances. Being a general method for Bayesian inference, collapsed Gibbs sampling is used to estimate the parameters $\bm{\theta}$ and $\bm{\phi}$ for LDA model by \cite{griffiths2004finding}. In the sampling procedure, the conditional distribution $p\left ( z_{m,n}=k|\bm{z}_{-m,n},w_{m,n}=v,\bm{w}_{-m,n},\alpha ,\beta  \right )$ is calculated where $\bm{z}_{-m,n}$ is the topic assignments for all tokens excluding $w_{m,n}$. Based on the Bayes rule and the property of conjugate priors, the probability that $\bm{z}_{m,n}$ is assigned to topic $k$ can be easily derived with the result given by

\begin{equation} \label{eq:LDA gibbs}
p\left ( z_{m,n}=k|\bm{z}_{-m,n},w_{m,n}=v,\bm{w}_{-m,n},\alpha ,\beta  \right )\propto \frac{n_{m,\left ( -m,n \right )}^{k} + \alpha }{n_{m,\left ( -m,n \right )}^{ \left ( \ast  \right )}+K\alpha }\frac{n_{k,\left ( -m,n \right )}^{v} + \beta  }{n_{k,\left ( -m,n \right )}^{ \left ( \ast  \right )}+V\beta }
\end{equation}

where $n_{m,\left ( -m,n \right )}^{k}$ denotes the number of tokens  assigned to topic $k$ in the $m$-th document excluding word $w_{m,n}$. $n_{m,\left ( -m,n \right )}^{ \left ( \ast  \right )}$ represents the total number of tokens in the $m$-th document excluding word $w_{m,n}$. $n_{k,\left ( -m,n \right )}^{v}$ is the number of tokens $v$ assigned to topic $k$ excluding word $w_{m,n}$. $n_{k,\left ( -m,n \right )}^{ \left ( \ast  \right )}$ is the total number of word tokens assigned to topic $k$ excluding word $w_{m,n}$.

The procedure of collapsed Gibbs sampling is summarized in Algorithm 1. Upon the convergence, we can estimate the parameters $\bm{\theta}$ and $\bm{\phi}$ as follows.
\begin{equation} \label{eq:update theta LDA}
    \widehat{\theta} _{m,k}= \frac{n_{m}^{k} + \alpha }{n_{m}^{ \left ( \ast  \right )}+K\alpha }
\end{equation}

\begin{equation} \label{eq:update phi LDA}
    \widehat{\phi } _{k,v}= \frac{n_{k}^{v} + \beta  }{n_{k}^{ \left ( \ast  \right )}+V\beta }
\end{equation}

\begin{table}
\centering
\begin{tabular}{lllp{7.4cm}}
\hline
{\textbf{Algorithm 1:}}  Procedure of collapsed Gibbs sampling for LDA \\ \hline
{\textbf{Input:}} topic number $K$, hyperparameters $\alpha$ and $\beta$, number of iterations $N_{iter}$ \\
{\textbf{Output:}} $\bm{\theta}$ and $\bm{\phi}$ \\
1. Randomly initialize the topic assignment $z_{m,n}$ for each word, and  update the count $n_{m}^{k}$, \\
\qquad  $n_{m}^{ \left ( \ast  \right )}$, $n_{k}^{v}$, $n_{k}^{ \left ( \ast  \right )}$ \\
2. For $iter=1$ to $N_{iter}$ \\
\qquad   -For each document $m\in \left [ 1,M \right ]$ \\
\qquad  \qquad  -For each word $w_{m,n}$ in document $m$ \\
\qquad  \qquad \qquad  Sample the topic $k$ using Equation (\ref{eq:LDA gibbs}) \\
\qquad  \qquad \qquad  Update the $n_{m}^{k}$,  $n_{m}^{ \left ( \ast  \right )}$, $n_{k}^{v}$, $n_{k}^{ \left ( \ast  \right )}$ \\
3. Estimate $\bm{\theta}$ using Equation (\ref{eq:update theta LDA}) and $\bm{\phi}$ using Equation (\ref{eq:update phi LDA}) \\
  \hline
\end{tabular}
\end{table}

\subsubsection{Collapsed variational Bayesian (CVB) inference for LDA}
Due to the stochastic nature, collapsed Gibbs sampling converges slowly, which makes it hard to scale to large corpus. While the standard variational Bayeisan, another popular parameter learning method, can overcome this limitation, it is efficient for large scale applications. However, the standard variational Bayeisan may lead to inaccurate estimates of the posterior comparing with collapsed Gibbs sampling (\cite{teh2007collapsed}). Collapsed variational Bayesian (CVB) inference for LDA combines the strengths of collapsed Gibbs sampling and standard variational Bayesian, thus offering a more effective and efficient approach to learn model parameters. The core idea is to simplify the evidence lower bound (ELBO) on $\textup{log}p\left ( \bm{w}|\alpha ,\beta  \right )$ by marginalizing out the parameters $\bm{\theta}$ and $\bm{\phi}$. Based on \cite{teh2007collapsed}, we can rewrite ELBO as follows.
\begin{equation} \label{eq:ELBO LDA}
    ELBO\left ( q\left ( \bm{z} \right ) \right )=\textup{E}_{q\left (  \bm{z}  \right )}\left [ \textup{log}p\left (  \bm{w} , \bm{z} |\alpha ,\beta  \right ) \right ]-\textup{E}_{q\left (  \bm{z}  \right )}\left [ \textup{log}q\left (  \bm{z}  \right ) \right ]
\end{equation}

where $p\left (  \bm{w} , \bm{z} |\alpha ,\beta  \right )$ is the marginal distribution over $\bm{w}$ and $\bm{z}$. $q\left (  \bm{z}  \right )$ is the variational distribution, which is approximation to true distribution $p(z)$. According to mean-field theory, $q\left (  \bm{z}  \right )$  is assumed to factorize over latent variables and thus  expressed as:

\begin{equation} \label{eq:q}
    q\left ( \bm{z} \right )=\prod_{m=1}^{M}\prod_{n=1}^{N_{m}}q\left ( z_{m,n}|\gamma _{m,n} \right )
\end{equation}
where $q\left ( z_{m,n}|\gamma _{m,n} \right )$ is restricted to be the Multinomial distribution parameterized by $\gamma _{m,n} $. The optimization goal is to maximize the ELBO, which is equivalent to minimize the difference between true posterior $p(z)$ and the variational distribution $q(z)$. The optimal variational parameters $\gamma _{m,n,k}=q\left ( z_{m,n}=k|\gamma _{m,n} \right )$ can computed by

\begin{equation} \label{eq:GAMMA}
\gamma _{m,n,k}=\frac{exp\left (\textup{E}_{q\left ( z_{-m,n} \right )}\textup{log}p\left ( z_{-m,n},z_{m,n}=k,w|\alpha ,\beta  \right ) \right )}{\sum_{k^{'}=1}^{K}exp\left (\textup{E}_{q\left ( z_{-m,n} \right )}\textup{log}p\left ( z_{-m,n},z_{m,n}=k^{'},w|\alpha ,\beta  \right )\right ) }
\end{equation}

However, it is computationally expensive for precisely implementing the CVB, since the algorithm need to compute expectations in Equation (\ref{eq:GAMMA}). Hence, \cite{teh2007collapsed} use a second-order Taylor expansion as an approximation to simplify the computation of expectation. To further reduce the computational costs, \cite{asuncion2009smoothing}  propose an approximate algorithm, Zero-Order Collapsed Variational Bayesian Inference Algorithm (CVB0), by using zero-order Taylor expansion. In CVB0, the update parameter $\gamma _{m,n,k}$ is updated as
\begin{equation} \label{eq:GAMMA cvb0}
\gamma _{m,n,k} \propto \left ( \widehat{n}_{m,\left ( -m,n \right )}^{k} + \alpha \right  )
\frac{\widehat{n}_{k,\left ( -m,n \right )}^{w_{m,n}} + \beta }{\widehat{n}_{k,\left ( -m,n \right )}^{\left ( \ast  \right )} + V\beta}
\end{equation}

where $\widehat{n}_{m,\left ( -m,n \right )}^{k}$ indicates the expected number of tokens in document $m$ assigned to topic $k$ excluding word $w_{m,n}$. $\widehat{n}_{k,\left ( -m,n \right )}^{w_{m,n}}$ denotes the expected number of tokens $w_{m,n}$ assigned to topic $k$ excluding word $w_{m,n}$. Formally, $\widehat{n}_{m,\left ( -m,n \right )}^{k}=\sum_{n^{'}\neq n}\gamma _{m,n^{'},k} $ and $\widehat{n}_{k,\left ( -m,n \right )}^{w_{m,n}}=\sum _{m^{'},n^{'},(m^{'},n^{'})\neq (m,n)}I[w_{m^{'},n^{'}}=w_{m,n}]\gamma _{m^{'},n^{'},k}$ where $I[\ast]$ is the indicator function.

The procedure of CVB0 is described in Algorithm 2. We estimate the parameters  $\bm{\theta}$ and $\bm{\phi}$  of the same form as Equation (\ref{eq:update theta LDA CVB}) and Equation (\ref{eq:update phi LDA CVB}).
\begin{equation} \label{eq:update theta LDA CVB}
    \widehat{\theta} _{m,k}= \frac{\widehat{n}_{m}^{k} + \alpha }{\widehat{n}_{m}^{ \left ( \ast  \right )}+K\alpha }
\end{equation}

\begin{equation} \label{eq:update phi LDA CVB}
    \widehat{\phi } _{k,v}= \frac{\widehat{n}_{k}^{v} + \beta  }{\widehat{n}_{k}^{ \left ( \ast  \right )}+V\beta }
\end{equation}

\begin{table}
\centering
\begin{tabular}{lllp{7.4cm}}
\hline
{\textbf{Algorithm 2:}}  Procedure of CVB0 for LDA \\ \hline
{\textbf{Input:}} topic number $K$, hyperparameters $\alpha$ and $\beta$, number of iterations $N_{iter}$ \\
{\textbf{Output:}} $\bm{\theta}$ and $\bm{\phi}$ \\
1. Randomly initialize the variational parameter $\gamma _{m,n,k}$, and update the value of $\widehat{n}_{m}^{k}$, \\
\qquad  $\widehat{n}_{m}^{ \left ( \ast  \right )}$, $\widehat{n}_{k}^{v}$, $\widehat{n}_{k}^{ \left ( \ast  \right )}$ \\
2. For $iter=1$ to $N_{iter}$ \\
\qquad   -For each document $m\in \left [ 1,M \right ]$ \\
\qquad  \qquad  -For each word $w_{m,n}$ in document $m$ \\
\qquad  \qquad \qquad -For each topic $k\in \left [ 1,K \right ]$ \\
\qquad  \qquad \qquad \qquad Update $\gamma _{m,n,k}$ using Equation (\ref{eq:GAMMA cvb0}) \\
\qquad  \qquad \qquad  \qquad Update the $\widehat{n}_{m}^{k}$,  $\widehat{n}_{m}^{ \left ( \ast  \right )}$, $\widehat{n}_{k}^{v}$, $\widehat{n}_{k}^{ \left ( \ast  \right )}$ \\
3. Estimate $\bm{\theta}$ using Equation (\ref{eq:update theta LDA CVB}) and $\bm{\phi}$ using Equation (\ref{eq:update phi LDA CVB}) \\
  \hline
\end{tabular}
\end{table}

\subsection{Sentence-LDA}
Sentence-LDA \citep{jo2011aspect,buschken2016sentence} is an extension of LDA which explictly considers the structure of
the text by assuming all words in a sentence share the same topic. Let $S_{m}$ denote the number of sentences in document $m$ and $N_{m,s}$ denote the total number of word tokens in sentence $s$ of document $m$, and the generative process of Sentence-LDA is described as follows.
\begin{enumerate}[1)]
    \item For each topic $k\in \left [ 1,K \right ]$
    \begin{enumerate}
        \item Draw $\bm{\phi _{k}}\sim Dirichlet\left ( \beta  \right )$
    \end{enumerate}
    \item For each document $m\in \left [ 1,M \right ]$
     \begin{enumerate}
        \item Draw $\bm{\theta _{m}}\sim Dirichlet\left ( \alpha  \right )$
        \item For each sentence $s\in \left [ 1,S_{m} \right ]$ in document $m$
        \begin{enumerate}
            \item Draw a topic $z_{m,s}\sim Multinomial\left ( \bm{\theta _{m}}  \right )$
            \item For each word $w \in \left [ 1,N_{m,s}\right ]$ in sentence $s$
            \begin{enumerate}
                \item  Draw a word $w \sim Multinomial\left ( \bm{\phi} _{z_{m,s}}  \right )$
            \end{enumerate}
        \end{enumerate}
    \end{enumerate}
\end{enumerate}

Consistent with \cite{jo2011aspect}, we also use the collapsed Gibbs sampling to estimate the parameters $\bm{\theta}$ and $\bm{\phi}$. The assignment of sentences to topics can be obtained by
\begin{equation} \label{eq:Sentence-LDA z}
    p\left ( z_{m,s}=k|\bm{z}_{-m,s},\bm{w},\alpha,\beta  \right )\propto \frac{n_{m,\left ( -m,s \right )}^{k}+\alpha  }{n_{m,\left ( -m,s \right )}^{ \left ( \ast  \right )}+K\alpha}\frac{\prod_{w\in s}\prod_{j=1}^{N_{m,s}^{w}}\left ( n_{k,-\left ( m,s \right )}^{w}+\beta +j-1 \right )}{\prod_{i=1}^{N_{m,s}}\left ( n_{k,\left (-m,s  \right )}^{ \left (\ast \right )} +V\beta +i-1\right )}
\end{equation}
where $n_{m,\left ( -m,s \right )}^{k}$  is the number of word tokens in document $m$ assigned to topic $k$ excluding the sentence $s$. $n_{m,\left ( -m,s \right )}^{ \left ( \ast  \right )}$ denotes the total number of word tokens in document $m$ excluding the sentence $s$. $N_{m,s}^{w}$ denotes the number of occurrences of word $w$ in the sentence $s$. $n_{k,-\left ( m,s \right )}^{w}$  is the number of tokens of word $w$ assigned to topic $k$  excluding the sentence $s$. $n_{k,\left (-m,s  \right )}^{ \left (\ast \right )}$ denotes the total number of word tokens assigned to topic $k$ excluding the sentence $s$.

We present the procedure of collapsed Gibbs sampling for Sentence-LDA in Algorithm 3. Specifically, the parameters $\bm{\theta}$ and $\bm{\phi}$ are given by Equation (\ref{eq:update theta SLDA}) and Equation (\ref{eq:update phi SLDA}).
\begin{equation} \label{eq:update theta SLDA}
    \widehat{\theta} _{m,k}= \frac{n_{m}^{k} + \alpha }{n_{m}^{ \left ( \ast  \right )}+K\alpha }
\end{equation}

\begin{equation} \label{eq:update phi SLDA}
    \widehat{\phi } _{k,v}= \frac{n_{k}^{v} + \beta  }{n_{k}^{ \left ( \ast  \right )}+V\beta }
\end{equation}

\begin{table}
\centering
\begin{tabular}{lllp{7.4cm}}
\hline
{\textbf{Algorithm 3:}} Procedure of collapsed Gibbs sampling for Sentence-LDA \\ \hline
{\textbf{Input:}} topic number $K$, hyperparameters $\alpha$ and $\beta$, number of iterations $N_{iter}$ \\
{\textbf{Output:}} $\bm{\theta}$ and $\bm{\phi}$ \\
1. Randomly initialize the topic assignment $z_{m,s}$ for each sentence, and  update the count $n_{m}^{k}$, \\
\qquad  $n_{m}^{ \left ( \ast  \right )}$, $n_{k}^{v}$, $n_{k}^{ \left ( \ast  \right )}$ \\
2. For $iter=1$ to $N_{iter}$ \\
\qquad   -For each document $m\in \left [ 1,M \right ]$ \\
\qquad  \qquad  -For each sentence  $s\in \left [ 1,S_{m} \right ]$ in document $m$ \\
\qquad  \qquad \qquad  Sample the topic $k$ using Equation (\ref{eq:Sentence-LDA z}) \\
\qquad  \qquad \qquad  Update the $n_{m}^{k}$,  $n_{m}^{ \left ( \ast  \right )}$, $n_{k}^{v}$, $n_{k}^{ \left ( \ast  \right )}$ \\
3. Estimate $\bm{\theta}$ using Equation (\ref{eq:update theta SLDA}) and $\bm{\phi}$ using Equation (\ref{eq:update phi SLDA}) \\
  \hline
\end{tabular}
\end{table}

\subsection{HDP}
HDP (hierarchical Dirichlet process) \citep{teh2005sharing} is a nonparametric Bayesian model on the basis of the Dirichlet process. LDA uses Dirichlet distribution to characterize the shades of membership to different topics, and assumes that the number of topics is fixed and finite. However, HDP assumes that there is an infinite number of topics, and allows the model to determine an optimal number of topics automatically.

The Dirichlet process $DP\left ( \gamma ,H \right )$ is in essence a measure on measures. It can be described by two parameters: a scaling (or concentration) parameter $\gamma > 0$ and a base measure $H$. To provide an explicit way to construct a Dirichlet process, \cite{sethuraman1994constructive} proposes the stick-breaking process. Specifically, a sample $G_0$ drawn from $DP\left ( \gamma ,H \right )$ can be defined as:
\begin{equation} \label{eq:dp}
    G_0=\sum_{k=1}^{\infty }\pi _{k}\delta _{\phi _{k}}
\end{equation}
where $\delta _{\phi _{k}}$ is a probability measure concentrated at $\phi _{k}$ which are independent and identically distributed draws from the base measure $H$. Parameters $\bm{\mathbf{\pi }}=\left \{ \pi _{k} \right \}_{k=1}^{\infty}$ denote a set of weights and satisfy $\sum_{k=1}^{\infty}\pi _{k}=1$. Each $\pi _{k}$ is defined on the basis of independent and identically distributed draws from $Beta(1,\gamma )$:
\begin{equation} \label{eq:stick-breaking}
\begin{split}
 &\pi _{k}=\pi _{k}{'}\prod_{l=1}^{k-1}\left ( 1- \pi _{l}{'}\right ) \\
 &\pi _{k}{'}|\gamma \sim Beta\left ( 1,\gamma  \right )
\end{split}
\end{equation}
For convenience and clarity, the above two steps in Equation (\ref{eq:stick-breaking}) are commonly written as $\bm{\mathbf{\pi }}\sim \mathrm{GEM}(\gamma)$ where $\mathrm{GEM} $ refers to Griffiths, Engen and McCloskey \citep{teh2005sharing}.

According to \cite{teh2005sharing}, HDP can be constructed via two-level stick-breaking representations. Particularly, in corpus level, the global measure $G_0$ follows the standard Dirichlet process, which is expressed as:
\begin{equation} \label{eq:corpus-level}
\begin{split}
  &G_0=\sum_{k=1}^{\infty }\pi _{k}\delta _{\phi _{k}} \\
  & \phi _{k}|H \sim H \\
  & \bm{\mathbf{\pi }}|\gamma =\left \{ \pi _{k} \right \}_{k=1}^{\infty} \sim GEM\left ( \gamma  \right )
\end{split}
\end{equation}
In document level, the random measure $G_m$ also follows standard Dirichlet process where the base measure is $G_0$. The details is given by:
\begin{equation} \label{eq:document-level}
\begin{split}
  &G_m=\sum_{k=1}^{\infty }\theta _{m,k}\delta _{\phi _{k}} \\
  & \theta _{m,k}=\theta _{m,k}^{'}\prod_{l=1}^{k-1}\left ( 1- \theta _{m,l}^{'}\right ) \\
  & \theta _{m,k}^{'}|\alpha _{0},\bm{\mathbf{\pi }}\sim Beta\left ( \alpha _{0}\pi _{k},\alpha _{0}\left ( 1-\sum_{l=1}^{k}\pi _{l} \right ) \right )
\end{split}
\end{equation}
HDP contains two additional steps for modeling the documents. First, it chooses a topic $z_{m,n}$ related to the word $w_{m,n}$, and then generates the word from $\phi _{z_{m,n}}$. The details are shown in Equation (\ref{eq:modeling the documents}).
\begin{equation} \label{eq:modeling the documents}
\begin{split}
& z_{m,n}|\bm{\theta} _{m} \sim Multinomial\left ( \bm{\theta} _{m} \right ) \\
& w_{m,n}|z_{m,n},\left \{ \phi_{k} \right \}_{k=1}^{\infty} \sim Multinomial\left ( \phi_{z_{m,n}} \right )
\end{split}
\end{equation}

For HDP, \cite{teh2005sharing} use Gibbs sampling on the basis of Chinese restaurant franchise to infer the topic assignment $z_{m,n}$. We first calculate the conditional density $f_{k}^{-m,n}$ of $w_{m,n}$ under topic $k$  given all data items except $w_{m,n}$:
\begin{equation} \label{eq:the conditional density }
\begin{aligned}
f_{k}^{-m,n}\left ( w_{m,n}=v \right )&=\frac{\int p\left ( w_{m,n}|\phi _{k}  \right )\prod_{m^{'},n^{'}\neq m,n;z_{m^{'},n^{'}}=k }p\left ( w_{m^{'},n^{'}}|\phi _{k} \right )p\left ( \phi _{k}|\beta  \right )d \phi _{k}}{\int\prod_{m^{'},n^{'}\neq m,n;z_{m^{'},n^{'}}=k }p\left ( w_{m^{'},n^{'}}|\phi _{k} \right )p\left ( \phi _{k}|\beta  \right )d \phi _{k}} \\
 &=\left\{
             \begin{array}{lr}
          \frac{n_{k}^{v}+\beta  }{n_{k}^{(\ast)}+V\beta },  \: \: \: \: \: if \,\, k \,\, exists &  \\

             \frac{ 1 }{V }  \: \: \: \: \: \: \: \: \: \: \: \: \: \: \: if \,\, k \,\,is \,\,new &  \\
             \end{array}
\right.
\end{aligned}
\end{equation}

Based on the idea Chinese restaurant franchise, we next calculate the probability of assigning a table $t$ to word $w_{m,n}$:
\begin{equation} \label{eq:assigning a table}
p\left ( t_{m,n}=t|\mathbf{t}_{-m,n},\mathbf{k} \right )\propto
\left\{
\begin{aligned}
n_{m,t,*}^{ -m,n}f_{k_{m,t}}^{-m,n}\left ( w_{m,n} \right )  \: \: \: \: \: \: \: \: \: \: \: \: \: \: \: \mathrm {if} \,\, t \,\,  \mathrm {previously} \,\,  \mathrm {used} \\
\alpha _{0}p\left ( w_{m,n}|\mathbf{t}_{-m,n},t_{m,n}=t^{new},\mathbf{k} \right ) \: \: \: \: \: \: \: \: \: \: \: \: \: \: \: \mathrm {if}  \,\, t = t^{new}\\
\end{aligned}
\right.
\end{equation}

where $n_{m,t,*}^{ -m,n}$ denotes the number of words in document $m$ that assign to table $t$, leaving out word $w_{m,n}$. Let $p\left ( w_{m,n}|\mathbf{t}_{-m,n},t_{m,n}=t^{new},\mathbf{k} \right )$
denote the likelihood for $ t_{m,n}=t^{new}$, which is calculated as
\begin{equation} \label{eq: likelihood }
p\left ( w_{m,n}|\mathbf{t}_{-m,n},t_{m,n}=t^{new},\mathbf{k} \right )=\sum_{k=1}^{K}\frac{m_{\ast k }}{m_{\ast \ast  }+\gamma }f_{k}^{-m,n}\left ( w_{m,n} \right )+\frac{\gamma }{m_{\ast \ast  }+\gamma}f_{k=k^{new}}^{-m,n}\left ( w_{m,n} \right )
\end{equation}
where $m_{\ast k }$ is the number of tables that belong to topic $k$. $m_{\ast \ast  }$ is the total number of tables.
If $t_{m,n}$ equals $t^{new}$, we sample a new topic $k_{mt^{new}}$ for the new table $t^{new}$:
\begin{equation} \label{eq: a new topic}
p\left ( k_{mt^{new}}=k|\mathbf{t},\mathbf{k}^{-mt^{new}} \right ) \propto
\left\{
\begin{aligned}
m_{*k}f_{k}^{-m,n}\left ( w^{_{w,n}} \right )  \: \: \: \: \: \: \: \: \: \: \: \: \: \: \: \mathrm {if} \,\, k \,\, \mathrm {exists}  \\
\gamma f_{k^{new}}^{-m,n}\left ( w_{m,n} \right ) \: \: \: \: \: \: \: \: \: \: \: \: \: \: \: \mathrm {if}  \,\, k = k^{new}\\
\end{aligned}
\right.
\end{equation}

We present the procedure of HDP learning in Algorithm 4. According to the topic assignments for each table, we compute $n_{m}^{k}$ by counting tokens in document $m$ assigned to topic $k$ and $n_{m}^{(\ast) }$ by counting tokens in document $m$.
Then, we estimate $\bm{\theta}$ and $\bm{\phi}$ by Equation (\ref{eq:update theta HDP}) and (\ref{eq:update phi hdp})
\begin{equation} \label{eq:update theta HDP}
    \widehat{\theta} _{m,k}= \frac{n_{m}^{k} + \alpha_{0} }{n_{m}^{ \left ( \ast  \right )}+K\alpha_{0} }
\end{equation}

\begin{equation} \label{eq:update phi hdp}
    \widehat{\phi } _{k,v}= \frac{n_{k}^{v} + \beta  }{n_{k}^{ \left ( \ast  \right )}+V\beta }
\end{equation}
\begin{table}
\centering
\begin{tabular}{lllp{7.4cm}}
\hline
{\textbf{Algorithm 4:}} Procedure of HDP learning \\ \hline
{\textbf{Input:}} initial topic number $K$, hyperparameters $\alpha_{0}$, $\beta$ and $\gamma$, number of iterations $N_{iter}$ \\
{\textbf{Output:}} $\bm{\theta}$ and $\bm{\phi}$ \\
1. Randomly initialize the table assignments $t_{m,n}$ for each word, topic assignment $k_{mt^{new}}$ for \\
\qquad  each new table, and  update the count $n_{m,t,\ast }$, $n_{k}^{v}$, $n_{k}^{ \left ( \ast  \right )}$, $m_{\ast k  }$ and $m_{\ast \ast  }$ \\
2. For $iter=1$ to $N_{iter}$ \\
\qquad   -For each document $m\in \left [ 1,M \right ]$ \\
\qquad  \qquad  -For each word $w_{m,n}$ in document $m$ \\
\qquad  \qquad \qquad  -Ensure the capacity so that new table and new topic can be chosen \\
\qquad  \qquad \qquad  -Sample the table $t$  using Equation (\ref{eq:assigning a table}) \\
\qquad  \qquad \qquad  -If $t=t^{new}$ \\
\qquad  \qquad \qquad \qquad  Sample the topic for this table using  Equation (\ref{eq: a new topic}) \\
\qquad  \qquad \qquad -Update the count $n_{m,t,\ast }$, $n_{k}^{v}$, $n_{k}^{ \left ( \ast  \right )}$, $m_{\ast k  }$ and $m_{\ast \ast  }$ \\
\qquad  \qquad \qquad - Remove the empty topic and table, and then re-arrange topic and table indices \\
3. Estimate $\bm{\theta}$ using Equation (\ref{eq:update theta HDP}) and $\bm{\phi}$ using Equation (\ref{eq:update phi hdp}) \\
  \hline
\end{tabular}
\end{table}

\subsection{DMM}
Short text clustering is an important task for NLP. Dirichlet Multinomial Mixture (DMM) model used in \cite{nigam2000text} is a popular generative model for short text clustering. This approach contains two assumptions: (1) each short document is generated from a cluster, and (2) each cluster is a multinomial distribution over words. The generative process of DMM is described as follows.
\begin{enumerate}[1)]
    \item For each cluster $k\in \left [ 1,K \right ]$
    \begin{enumerate}
        \item Draw $\bm{\phi _{k}}\sim Dirichlet\left ( \beta  \right )$
    \end{enumerate}
    \item Draw the global cluster distribution $\bm{\theta}\sim Dirichlet\left ( \alpha  \right )$
    \item For each document $m\in \left [ 1,M \right ]$
     \begin{enumerate}
        \item Draw the cluster $z_{m} \sim Multinomial\left ( \bm{\theta}  \right )$
        \item For each word $w_{m,n}$ in document $m$
        \begin{enumerate}
            \item Draw $w_{m,n} \sim Multinomial\left ( \bm{\phi _{z_{m}}}  \right )$
        \end{enumerate}
    \end{enumerate}
\end{enumerate}
Define $\theta _{k}=p\left ( z=k|\bm{\theta}  \right )$, then $\bm{\theta} =\left \{ \theta _{k} \right \}_{k=1}^{K}$ represents
the weight of the mixture clusters in the corpus satisfying $\sum_{k=1}^{K}\theta _{k}=1$.

\cite{yin2014dirichlet} propose a collapsed Gibbs Sampling algorithm for DMM. Using the Bayes Rule, the allocation of short document to cluster can be derived based on the conditional distribution $p\left ( z_{m}|\bm{z}_{-m},\bm{d},\alpha ,\beta  \right )$.
\begin{equation} \label{eq:DMM}
\begin{split}
p\left ( z_{m}|\bm{z}_{-m},\bm{d},\alpha ,\beta  \right )&=\frac{p\left ( \bm{d},\bm{z}|\alpha ,\beta  \right )}{p\left ( \bm{d},\bm{z}_{-m}|\alpha ,\beta  \right )} \\
& \propto \frac{p\left ( \bm{d},\bm{z}|\alpha ,\beta  \right )}{p\left ( \bm{d}_{-m},\bm{z}_{-m}|\alpha ,\beta  \right )} \\
& = \frac{n_{k,-m}+\alpha }{M-1+K\alpha }\frac{\prod _{w\in m}\prod_{j=1}^{N_{m}^{w}}\left ( n_{k,-m}^{w}+\beta +j+1 \right )}{\prod_{i=1}^{N_{m}}\left ( n_{k,-m}^{\ast } + V\beta +i-1 \right )}
\end{split}
\end{equation}

where $\bm{d}$ represents a set of documents in the corpus. $\bm{z}_{-m}$ denotes the cluster labels of each document excluding document $m$. $ n_{k,-m}$ denotes the number of documents assigned to cluster $k$ excluding document $m$. $N_{m}^{w}$ is the number of occurrences of word $w$  in document $m$. $N_{m}$ is the number of words in document $m$. $n_{k,-m}^{w}$ denotes the number of words $w$ assigned to cluster $k$ excluding the document $m$. $n_{k,-m}^{\ast }$ is the total number of words assigned to cluster $k$ excluding the document $m$.

We present the procedure of collapsed Gibbs sampling for DMM in Algorithm 5. And the parameters $\bm{\theta}$ and $\bm{\phi}$ are given by Equation (\ref{eq:theta for DMM}) and Equation (\ref{eq:phi for DMM}).
\begin{equation} \label{eq:theta for DMM}
\widehat{\theta} _{k}=\frac{n_{k}+\alpha }{M+K\alpha }
\end{equation}

\begin{equation} \label{eq:phi for DMM}
\widehat{\phi } _{k,v}= \frac{n_{k}^{v} + \beta  }{n_{k}^{ \left ( \ast  \right )}+V\beta }
\end{equation}

\begin{table}
\centering
\begin{tabular}{lllp{7.4cm}}
\hline
{\textbf{Algorithm 5:}} Procedure of collapsed Gibbs sampling for DMM \\ \hline
{\textbf{Input:}} Cluster number $K$, hyperparameters $\alpha$, and $\beta$, number of iterations $N_{iter}$ \\
{\textbf{Output:}} $\bm{\theta}$ and $\bm{\phi}$ \\
1. Randomly initialize the cluster assignments $z_{m}$ for each document, and  update \\
\qquad the count $n_{k}$, $n_{k}^{v}$ and $n_{k}^{\left ( \ast  \right )}$ \\
2. For $iter=1$ to $N_{iter}$ \\
\qquad   -For each document $m\in \left [ 1,M \right ]$ \\
\qquad  \qquad  Sample the cluster k using quation (\ref{eq:DMM})  \\
\qquad  \qquad  Update the count $n_{k}$, $n_{k}^{v}$ and $n_{k}^{\left ( \ast  \right )}$ \\
3. Estimate $\bm{\theta}$ using Equation (\ref{eq:theta for DMM}) and $\bm{\phi}$ using Equation (\ref{eq:phi for DMM}) \\
  \hline
\end{tabular}
\end{table}

\subsection{DPMM}
Dirichlet Process Multinomial Mixture (DPMM) can be considered as an infinite extension of the DMM model. In DMM model, the cluster number $K$ is fixed and finite. In contrast, DPMM can automatically determine the appropriate number of clusters on the basis of  the properties of
the Dirichlet Process. The generative process of the DPMM is described as follows:
\begin{enumerate}[1)]
    \item For each cluster $k=1,2,\cdots,\infty $
    \begin{enumerate}
        \item Draw $\bm{\phi _{k}}\sim Dirichlet\left ( \beta  \right )$
    \end{enumerate}
    \item Draw the global cluster distribution $\bm{\theta}\sim GEM\left ( \alpha  \right )$
    \item For each document $m\in \left [ 1,M \right ]$
     \begin{enumerate}
        \item Draw the cluster $z_{m} \sim Multinomial\left ( \bm{\theta}  \right )$
        \item For each word $w_{m,n}$ in document $m$
        \begin{enumerate}
            \item Draw $w_{m,n} \sim Multinomial\left ( \bm{\phi _{z_{m}}}  \right )$
        \end{enumerate}
    \end{enumerate}
\end{enumerate}

Suggested by the the collapsed Gibbs sampling algorithm for the DPMM model proposed by \cite{yin2016model}, we sample the cluster $z_{m}$ for each document according to the conditional distribution
$p\left ( z_{m}|\bm{z}_{-m},\bm{d},\alpha ,\beta  \right )$ shown in Equation (\ref{eq:DPMM conditional distribution}).
\begin{equation} \label{eq:DPMM conditional distribution}
p\left ( z_{m}|\bm{z}_{-m},\bm{d},\alpha ,\beta  \right )\propto p\left ( z_{m}=k|\bm{z}_{-m},\alpha  \right )p\left ( m| z_{m}=k,\beta \right )
\end{equation}
The first term of the right-hand side of Equation (\ref{eq:DPMM conditional distribution}) is derived as follows:
\begin{equation} \label{eq:Equation first}
p\left ( z_{m}=k|\bm{z}_{-m},\alpha  \right )=\int p\left ( \bm{\theta} |\bm{z}_{-m},\alpha  \right )p\left ( z_{m}=k|\bm{\theta}  \right )d\bm{\theta} =\frac{n_{k,-m}+\alpha /K}{M-1+\alpha }
\end{equation}
Meanwhile, the derivation of second term is given by:
\begin{equation} \label{eq:Equation second}
\begin{split}
p\left ( m| z_{m}=k,\beta \right )&=\int p\left ( \bm{\phi} _{k}|\bm{d}_{-m},\beta  \right )p\left ( \bm{\phi} _{k},z_{m}=k \right )d\bm{\phi} _{k} \\
&=\frac{\prod _{w\in m}\prod_{j=1}^{N_{m}^{w}}\left ( n_{k,-m}^{w}+\beta +j+1 \right )}{\prod_{i=1}^{N_{m}}\left ( n_{k,-m}^{\ast } + V\beta +i-1 \right )}
\end{split}
\end{equation}
Hence, the probability that document $m$ is assigned to cluster $k$ given other information is given by:
\begin{equation} \label{eq:DPMM probability}
p\left ( z_{m}|\bm{z}_{-m},\bm{d},\alpha ,\beta  \right )\propto \frac{n_{k,-m}+\alpha /K}{M-1+\alpha }\frac{\prod _{w\in m}\prod_{j=1}^{N_{m}^{w}}\left ( n_{k,-m}^{w}+\beta +j+1 \right )}{\prod_{i=1}^{N_{m}}\left ( n_{k,-m}^{\ast } + V\beta +i-1 \right )}
\end{equation}
Let $K$ go to infinity ($\infty$), we can obtain the probability of document $m$ assigned to an existing cluster $k$ as follows:
\begin{equation} \label{eq:DPMM infinity}
p\left ( z_{m}|\bm{z}_{-m},\bm{d},\alpha ,\beta  \right )\propto \frac{n_{k,-m}}{M-1+\alpha }\frac{\prod _{w\in m}\prod_{j=1}^{N_{m}^{w}}\left ( n_{k,-m}^{w}+\beta +j+1 \right )}{\prod_{i=1}^{N_{m}}\left ( n_{k,-m}^{\ast } + V\beta +i-1 \right )}
\end{equation}
Denoting a new cluster as $k^{new}$, we obtain the probability of document $m$ being assigned to cluster $k^{new}$ as:
\begin{equation} \label{eq:new cluster}
p\left ( z_{m}|\bm{z}_{-m},\bm{d},\alpha ,\beta  \right )\propto \frac{\alpha}{M-1+\alpha }\frac{\prod _{w\in m}\prod_{j=1}^{N_{m}^{w}}\left (\beta +j+1 \right )}{\prod_{i=1}^{N_{m}}\left ( V\beta +i-1 \right )}
\end{equation}
Finally, we combine Equation (\ref{eq:DPMM infinity}) and Equation (\ref{eq:new cluster}), and obtain:
\begin{equation} \label{eq:Finally}
p\left ( z_{m}|\bm{z}_{-m},\bm{d},\alpha ,\beta  \right )\propto
\left\{
\begin{aligned}
\frac{n_{k,-m}}{M-1+\alpha }\frac{\prod _{w\in m}\prod_{j=1}^{N_{m}^{w}}\left ( n_{k,-m}^{w}+\beta +j+1 \right )}{\prod_{i=1}^{N_{m}}\left ( n_{k,-m}^{\ast } + V\beta +i-1 \right )}  \: \: \: \: \: \: \: \: \: \: \: \: \: \: \: \mathrm {if} \,\, k \,\,  \mathrm {exists} \\
\frac{\alpha}{M-1+\alpha }\frac{\prod _{w\in m}\prod_{j=1}^{N_{m}^{w}}\left (\beta +j+1 \right )}{\prod_{i=1}^{N_{m}}\left ( V\beta +i-1 \right )} \: \: \: \: \: \: \: \: \: \: \: \: \: \: \: \mathrm {if}  \,\, k = k^{new}\\
\end{aligned}
\right.
\end{equation}
We present the procedure of collapsed Gibbs sampling for DPMM in Algorithm 6. Similar to DMM, we can estimate the parameters $\bm{\theta}$ and $\bm{\phi}$ by Equation (\ref{eq:theta for DPMM}) and Equation (\ref{eq:phi for DPMM}).
\begin{equation} \label{eq:theta for DPMM}
\widehat{\theta} _{k}=\frac{n_{k}+\alpha }{M+K\alpha }
\end{equation}
\begin{equation} \label{eq:phi for DPMM}
\widehat{\phi } _{k,v}= \frac{n_{k}^{v} + \beta  }{n_{k}^{ \left ( \ast  \right )}+V\beta }
\end{equation}
\begin{table}
\centering
\begin{tabular}{lllp{7.4cm}}
\hline
{\textbf{Algorithm 6:}} Procedure of collapsed Gibbs sampling for DPMM \\ \hline
{\textbf{Input:}} initial cluster number $K$, hyperparameters $\alpha$, and $\beta$, number of iterations $N_{iter}$ \\
{\textbf{Output:}} $\bm{\theta}$ and $\bm{\phi}$ \\
1. Randomly initialize the cluster assignments $z_{m}$ for each document, and  update \\
\qquad the count $n_{k}$, $n_{k}^{v}$ and $n_{k}^{\left ( \ast  \right )}$ \\
2. For $iter=1$ to $N_{iter}$ \\
\qquad   -For each document $m\in \left [ 1,M \right ]$ \\
\qquad  \qquad -Record the current cluster of $m$ as $k$, and update the count $n_{k,-m}$, $n_{k,-m}^{w}$ and $n_{k,-m}^{\left ( \ast \right ) }$ \\
\qquad  \qquad -If $n_{k}=0$  \\
\qquad  \qquad \qquad //Remove the inactive cluster \\
\qquad  \qquad \qquad $K=K-1$  \\
\qquad  \qquad \qquad Re-arrange cluster indices \\
\qquad  \qquad  -Sample the cluster k using Equation (\ref{eq:Finally})  \\
\qquad  \qquad  -If $k=k^{new}$  \\
\qquad  \qquad \qquad //A new cluster \\
\qquad  \qquad \qquad $K=K+1$   \\
\qquad  \qquad \qquad Initialize the count $n_{k}$, $n_{k}^{v}$ and $n_{k}^{\left ( \ast  \right )}$ as zero \\
\qquad  \qquad  -Update the count $n_{k}$, $n_{k}^{v}$ and $n_{k}^{\left ( \ast  \right )}$ \\
3. Estimate $\bm{\theta}$ using Equation (\ref{eq:theta for DPMM}) and $\bm{\phi}$ using Equation (\ref{eq:phi for DPMM}) \\
  \hline
\end{tabular}
\end{table}

\subsection{PTM}
Due to the sparsity of word co-occurrences in short documents,
LDA struggles to obtain satisfactory performances. To solve this problem,
Pseudo-document-based Topic Model (PTM) is proposed by \cite{zuo2016topic} where pseudo documents are introduced
for indirect aggregation of short texts. In PTM, a
large of short documents is transformed into relatively few pseudo documents (long text) to avoid data sparsity, which could improve the topic learning.

We assume that there are $P$ pseudo documents and each one is denoted as $l\in \left [ 1,P \right ]$. Let $\psi$ denote a multinomial distribution which models the distribution of short documents over pseudo documents. In PTM, each short document is assigned to one pseudo document. For generating each word in a short document, PTM first sample a topic $z$ from topic distribution $\bm{\theta}$ over pseudo document, and then sample a word from the topic-word distribution $\bm{\phi}_{z}$. The generative process of the PTM is described as follows:
\begin{enumerate}[1)]
    \item Draw $\varphi \sim Dirichlet\left ( \lambda  \right )$
    \item For each topic $k\in \left [ 1,K \right ]$
    \begin{enumerate}
        \item Draw $\bm{\phi} _{k}\sim Dirichlet\left ( \beta  \right )$
    \end{enumerate}
    \item For each pseudo document $l\in \left [ 1,P \right ]$
    \begin{enumerate}
        \item Draw $\bm{\theta} _{l} \sim Dirichlet\left ( \alpha  \right )$
    \end{enumerate}
    \item For each short document $m\in \left [ 1,M \right ]$
     \begin{enumerate}
        \item Draw a pseudo document $l_{m} \sim Multinomial\left ( \varphi  \right )$
        \item For each word $w_{m,n}$ in document $m$
        \begin{enumerate}
             \item Draw the topic $z_{m,n} \sim Multinomial\left ( \bm{\theta}_{l_{m}}  \right )$
            \item Draw $w_{m,n} \sim Multinomial\left ( \bm{\phi} _{z_{m,n}}  \right )$
        \end{enumerate}
    \end{enumerate}
\end{enumerate}
Same as \cite{zuo2016topic}, we use the collapsed Gibbs sampling algorithm for model estimation. In PTM, there are two latent variables needed to iterate by the sampling algorithm: pseudo document assignment $l$ and topic assignment $z$. Conditioned on all the other documents, the probability
of a short document $l_m$ being assigned to a pseudo document $l$ is written as:
\begin{equation} \label{eq:document assignment}
p\left ( l_{m}=l|rest \right )\propto \frac{n_{l,-m}+\lambda }{M-1+P\lambda }\frac{\prod _{k\in m}\prod_{j=1}^{N_{m}^{k}} \left ( N_{l,-m}^{k}+\alpha +j-1 \right )}{\prod_{i=1}^{N_{m}}\left ( N_{l,-m}^{\left ( \ast  \right ) }+K\alpha +i-1 \right )}
\end{equation}
where $n_{l,-m}$ is the number of short documents assigned to the pseudo document $l$ excluding document $m$. $N_{m}^{k}$ is the number of words assigned to topic $k$ in document $m$. $N_{l,-m}^{k}$ denotes the number of words assigned to topic $k$  in pseudo document $l$ excluding document $m$. $N_{l,-m}^{\left ( \ast  \right ) }$ denotes is the total number of words in pseudo document $l$ excluding document $m$. $N_{m}$ is the number of words in document $m$.

We can sample the topic $z_{m,n}$ for each word according to the conditional probability of $z_{m,n}=k$, which is given by:
\begin{equation} \label{eq:topic assignment}
p\left ( z_{m,n}=k|w_{m,n}=v,rest \right )\propto \frac{N_{l,\left ( -m,n \right )}^{k}+\alpha }{N_{l,\left ( -m,n \right )}^{\ast }+K\alpha }\frac{n_{k,(-m,n)}^{v} + \beta  }{n_{k,(-m,n)}^{ \left ( \ast  \right )}+V\beta }
\end{equation}
Algorithm 7 shows the procedure of collapsed Gibbs sampling for PTM. Note that PTM learn the topic distribution $\bm{\theta} _{l}$ for
each pseudo document $l$. However, we can count $n_{m}^{k}$ and $n_{m}^{(\ast )}$ according to topic assignment $z_{m,n}$.
Then we obtain topic distribution $\bm{\theta} _{m}$ for each short document as follows:
\begin{equation} \label{eq:theta for PTM}
\widehat{\theta} _{m,k}= \frac{n_{m}^{k} + \alpha }{n_{m}^{ \left ( \ast  \right )}+K\alpha }
\end{equation}
From Equation (\ref{eq:phi for PTM1}), we can estimate the parameter $\bm{\phi}$.
\begin{equation} \label{eq:phi for PTM1}
\widehat{\phi } _{k,v}= \frac{n_{k}^{v} + \beta  }{n_{k}^{ \left ( \ast  \right )}+V\beta }
\end{equation}
\begin{table}
\centering
\begin{tabular}{lllp{7.4cm}}
\hline
{\textbf{Algorithm 7:}} Procedure of collapsed Gibbs sampling for PTM \\ \hline
{\textbf{Input:}} topic number $K$, number of pseudo document $P$, hyperparameters $\alpha$, $\beta$ and $\gamma$, \\
\qquad number of iterations $N_{iter}$ \\
{\textbf{Output:}} $\bm{\theta}$ and $\bm{\phi}$ \\
1. Randomly initialize the pseudo document assignment $l_{m}$ for each document;  \\
\qquad randomly initialize the topic assignments $z_(m,n)$ for each word; \\
\qquad update the count $n_l$, $N_l^k$, $N_l^{(*)}$,$n_k^v$  and $n_k^((*) )$ \\
2. For $iter=1$ to $N_{iter}$ \\
\qquad   -For each document $m\in \left [ 1,M \right ]$ \\
\qquad  \qquad Sample the pseudo document assignment $l$ using Equation (\ref{eq:document assignment}) \\
\qquad  \qquad Update the count $n_l$, $N_l^k$ and $N_l^{(*)}$  \\
\qquad  -For each document $m\in \left [ 1,M \right ]$ \\
\qquad  \qquad -For each word  $w_{m,n}$ in document $m$  \\
\qquad  \qquad \qquad Sample the topic $k$ using Equation (\ref{eq:topic assignment}) \\
\qquad  \qquad \qquad Update the count $N_l^k$, $N_l^{(*)}$, $n_k^v$  and $n_k^{(*)}$  \\
3. Estimate $\bm{\theta}$ using Equation (\ref{eq:theta for PTM}) and $\bm{\phi}$ using Equation (\ref{eq:phi for PTM1}) \\
  \hline
\end{tabular}
\end{table}

\subsection{BTM}
To cope with the sparsity of word co-occurrence patterns in short text, BTM is proposed by \cite{cheng2014btm,yan2013biterm}.
Unlike LDA and DMM, BTM assumes that a biterm is an unordered word-pair generated from a short text, and the two words in the biterm share the same topic. In addition, each topic for the biterm is drawn from a global topic distribution over the corpus.

In BTM, the first step is to construct biterms on the basis of short texts and a fixed-size window. For example, we can construct three biterms from a document with three distinct words:
\begin{equation} \label{eq:phi for PTM}
(w_1,w_2,w_2 )\rightarrow {(w_1,w_2 ),(w_2,w_3 ),(w_1,w_3 )}
\end{equation}
The biterm construction process is a single scan over all documents, resulting in a biterm set for the corpus.
We use $\bm{B}=\left \{ b_{i} \right \}_{i=1}^{N_{B}}$ to denote all biterms ($b_{i}=\left \{ w_{i,1},w_{i,2} \right \}$), $N_{B}$ to
 denote the number of biterms, $\bm{\theta} =\left \{ \theta _{k} \right \}_{k=1}^{K}$ where $\theta _{k}=p\left ( z=k|\bm{\theta}  \right )$ to
  denote the weight of the mixture clusters satisfying $\sum_{k=1}^{K}\theta _{k}=1$.
The generative process of the BTM is described as follows:
\begin{enumerate}[1)]
    \item For each topic $k\in \left [ 1,K \right ]$
    \begin{enumerate}
        \item Draw $\bm{\phi} _{k}\sim Dirichlet\left ( \beta  \right )$
    \end{enumerate}
    \item Draw the global topic distribution $\bm{\theta}\sim Dirichlet\left ( \alpha  \right )$
    \item For each biterm $b_i \in \bm{B}$
    \begin{enumerate}
        \item Draw the topic $z_i\sim Multinomial(\bm{\theta})$
        \item $w_{i,1},w_{i,2} \sim Multinomial(\bm{\phi}_{z_i} )$
    \end{enumerate}
\end{enumerate}

Same as \cite{cheng2014btm}, we use the collapsed Gibbs sampling algorithm to estimate parameters. The topic $z_i$ for
each biterm $b_i$ is drawn from the conditional probability $p(z_i=k|\bm{z}_{-i},\bm{B})$,
which can be derived elegantly using conjugate priors and Bayes rule as follows:
\begin{equation} \label{eq:conditional probability for BTM}
\begin{split}
p\left ( z_i|\bm{z}_{-i},B \right )&=\frac{p\left ( \bm{z},\bm{B} \right )}{p\left ( \bm{z}_{-i},\bm{B} \right )}\propto \frac{p\left ( \bm{B}|\bm{z} \right )p\left ( \bm{z} \right )}{p\left ( \bm{B}_{-i}|z_{i} \right )p\left ( \bm{z}_{-i} \right )} \\
&=\frac{n_{k,-i}+\alpha }{N_{B}-1+K\alpha }\frac{\left ( n_{k,-i}^{w_{i,1}}+\beta  \right )\left ( n_{k,-i}^{w_{i,2}}+\beta  \right )}{\left ( n_{k,-i}^{(\ast )}+V\beta +1  \right )  \left ( n_{k,-i}^{(\ast )}+V\beta  \right )}
\end{split}
\end{equation}
where $\bm{z}_{-i}$ is the topic assignments for all biterms excluding biterm $b_i$. $n_{k,-i}$ denotes the number of biterms assigned to topic $k$ excluding biterm $b_i$. $n_{k,-i}^v$ denotes the number of times word $v$ assigned to topic $k$ excluding biterm $b_i$.

Algorithm 8 summarizes the procedure of collapsed Gibbs sampling for BTM. And we can estimate the parameters $\bm{\theta}$ and $\bm{\phi}$ by Equation (\ref{eq:theta for BTM})  and Equation  Equation (\ref{eq:phi for BTM}).
\begin{equation} \label{eq:theta for BTM}
\widehat{\theta}_{k}=\frac{n_{k}+\alpha }{N_{B}+K\alpha }
\end{equation}
\begin{equation} \label{eq:phi for BTM}
\widehat{\phi } _{k,v}= \frac{n_{k}^{v} + \beta  }{n_{k}^{ \left ( \ast  \right )}+V\beta }
\end{equation}
For the estimatation of document-topic distribution, denoted as $p(k|m)$, we use the Bayes rule
 according to the topic assignment of biterms, which gives:
\begin{equation} \label{eq:document topic distribution}
\begin{split}
p\left ( k|m \right )&=\sum_{i=1}^{N_{m}}p\left ( k,b_{i}^{(m)}|m \right )=\sum_{i=1}^{N_{m}}p\left ( k|b_{i}^{m},m \right )p\left ( b_{i}^{(m)}|m \right ) \\
&=\sum_{i=1}^{N_{m}}\frac{ \hat{\theta}_{k}\hat{\phi}_{k,w_{i,1}^{(m)}}\hat{\phi}_{k,w_{i,2}^{(m)}}  }{\sum_{k^{'}=1}^{K}\hat{\theta}_{k}\hat{\phi}_{k,w_{i,1}^{(m)}}\hat{\phi}_{k,w_{i,2}^{(m)}} }\frac{n\left ( b_{i}^{(m)} \right )}{N_{m}}
\end{split}
\end{equation}
where $N_m$ is the number of biterms in document $m$. $b_{i}^{(m)}=\left \{ w_{i,1}^{(m)},w_{i,2}^{(m)} \right \} $ is a biterm of document $m$. $n(b_i^{(m)} )$ is the frequency of biterm $b_i^{(m)}$ in document $m$.
\begin{table}
\centering
\begin{tabular}{lllp{7.4cm}}
\hline
{\textbf{Algorithm 8:}} Procedure of collapsed Gibbs sampling for BTM \\ \hline
{\textbf{Input:}} topic number $K$, hyperparameters $\alpha$ and $\beta$, number of iterations $N_{iter}$ \\
{\textbf{Output:}} $\bm{\theta}$, $\bm{\phi}$ and $p(k|m)$ \\
1. Construct biterms on the basis of short texts and a fixed-size window  \\
2. Randomly initialize the topic assignment $z_i$ for each biterm, and update the count $n_k$, $n_k^v$  , $n_k^{(*)}$  \\
3. For $iter=1$ to $N_{iter}$ \\
\qquad   -For each biterm $b_{i}\in \bm{B}$ \\
\qquad  \qquad Sample the topic $k$ using Equation (\ref{eq:conditional probability for BTM}) \\
\qquad  \qquad Update the count $n_k$, $n_k^v$  , $n_k^{(*)}$  \\
4. Estimate $\bm{\theta}$ using Equation (\ref{eq:theta for BTM}) and $\bm{\phi}$ using Equation (\ref{eq:phi for BTM}) \\
5. Estimate $p(k|m)$ using Equation (\ref{eq:document topic distribution}) \\
  \hline
\end{tabular}
\end{table}

\subsection{ATM}
To capture authorship information, a critical aspect for documents such as academic publications, Author-topic model (ATM) \citep{rosen2004author} is
 proposed with the underlying assuption that each author is related to a multinomial distribution over topics while each topic is characterised by a
 multinomial distribution over words. For documents with multiple authors, the distribution over topics is an admixture of the distributions
 corresponding to the authors.
In this way, ATM can discover salient latent topics as well as interesting patterns for each author.

In ATM, a document is generated via the following process. First, an author is randomly selected for each word. Second,
a topic is chosen from the author-topic multinomial distribution associated with the selected author. Finally, a word is drawn from
the topic-word multinomial distribution conditioned on the chosen topic. Let $\bm{a}_m$ denote the authors of document $m$ and $x_{m,n}$
denote the author assignment for the word $w_{m,n}$. The generative process is detailed as follows:
\begin{enumerate}[1)]
    \item For each author $a\in \left [ 1,A \right ]$
    \begin{enumerate}
        \item Draw $\bm{\theta}_{a} \sim Dirichlet\left ( \alpha  \right )$
    \end{enumerate}
    \item For each topic $k\in \left [ 1,K \right ]$
    \begin{enumerate}
        \item Draw $\bm{\phi} _{k}\sim Dirichlet\left ( \beta  \right )$
    \end{enumerate}
    \item For each document $m \in \left [ 1,M \right ]$
    \begin{enumerate}
        \item Given the authors $\bm{a}_m$ of document $m$
        \item For each word  $w_{m,n}$ in document $m$
        \begin{enumerate}
            \item Draw an author $x_{m,n} \sim Uniform(\bm{a}_m )$
            \item Draw the topic $z_{m,n}\sim Multinomial(\bm{\theta}_{x_{m,n}})$
            \item Draw the word $w_{m,n}\sim Multinomial(\bm{\phi}_{z_{m,n}}) $
        \end{enumerate}
     \end{enumerate}
\end{enumerate}
Same as \cite{rosen2004author}, we apply the collapsed Gibbs sampling algorithm to estimate the parameters. For each word $w_{m,n}$,
the associated author and topic are sampled simultaneously according to the conditional probability given by:
\begin{equation} \label{eq: conditional probability for ATM}
p\left ( x_{m,n},z_{m,n}|w_{m,n}=v,\bm{z}_{-m,n},\bm{x}_{-m,n},\bm{w}_{-m,n},\bm{A},\alpha,\beta  \right )\propto \frac{n_{a,\left ( -m,n \right )}^{k}+\alpha }{n_{a,\left ( -m,n \right )}^{\left ( \ast  \right )}+ K\alpha}\frac{n_{k,\left ( -m,n \right )}^{v} + \beta  }{n_{k,\left ( -m,n \right )}^{ \left ( \ast  \right )}+V\beta }
\end{equation}
where $n_{a,(-m,n)}^k$ is the number of times author $a$ assigned to topic $k$ excluding the assignment of word $w_{m,n}$. $n_{a,(-m,n)}^{(*) }$ is the total number of the author $a$ excluding the assignment of word $w_{m,n}$.

Algorithm 9 summarizes the procedure of collapsed Gibbs sampling for ATM. We estimate the parameters $\bm{\theta}$ and $\bm{\phi}$ by Equation  (\ref{eq:theta for ATM}) and Equation (\ref{eq:phi for ATM}).
\begin{equation} \label{eq:theta for ATM}
\widehat{\theta}_{k,a}=\frac{n_{a}^{k}+\alpha }{n_{a}^{(\ast )}+K\alpha }
\end{equation}
\begin{equation} \label{eq:phi for ATM}
\widehat{\phi } _{k,v}= \frac{n_{k}^{v} + \beta  }{n_{k}^{ \left ( \ast  \right )}+V\beta }
\end{equation}
\begin{table}
\centering
\begin{tabular}{lllp{7.4cm}}
\hline
{\textbf{Algorithm 9:}} Procedure of collapsed Gibbs sampling for ATM \\ \hline
{\textbf{Input:}} topic number $K$, hyperparameters $\alpha$ and $\beta$, number of iterations $N_{iter}$ \\
{\textbf{Output:}} $\bm{\theta}$, $\bm{\phi}$ \\
1. Randomly initialize the topic assignment $z_{m,n}$ and the author assignment $a$ for each word;  \\
\qquad  update the count $n_a^k$, $n_a^{(*)}$, $n_k^v$, $n_k^{(*)}$ \\
2. For $iter=1$ to $N_{iter}$ \\
\qquad   -For each document $m \in [1,M]$ \\
\qquad \qquad -For each word  $w_{m,n}$ in document $m$  \\
\qquad  \qquad \qquad Sample the topic $k$ and the author $a$ using Equation (\ref{eq: conditional probability for ATM}) \\
\qquad  \qquad \qquad Update the count $n_a^k$, $n_a^{(*)}$, $n_k^v$, $n_k^{(*)}$   \\
3. Estimate $\bm{\theta}$ using Equation (\ref{eq:theta for ATM}) and $\bm{\phi}$ using Equation (\ref{eq:phi for ATM}) \\
  \hline
\end{tabular}
\end{table}

\subsection{Link LDA}
Given the nature of linkage in documents such as blogs and research literature, Link LDA \citep{erosheva2004mixed} is proposed as a
unified framework to discover document topics and to analyze author community at the same time. Particularly, following the intuition that
the content of a document is closely associated with its citations, the Link LDA assumes that in each document the topic assignments for words and
for citations are drawn from the same topic distribution (i.e., the topic distribution of the corresponding document). Link LDA generates a document
and its citations via a two-stage process. The first stage is the generation of document. Similar to LDA, it first picks a topic from multinomial distribution over topics. Then, the word is drawn from the multinomial distribution over words related to the topic.  The second stage is the generation of
citations. It also picks a topic from multinomial distribution over topics. Then, the citation is drawn from the multinomial distribution over citations related to the topic.

We use $l_{m,n}$ to refer to the $l$-th citation link for document $m$, $L$ to refer to the number of unique links in the vocabulary,
$z_{m,n}$ to refer to the topic assignment for the word $w_{m,n}$, $x_{m,n}$ to refer to the topic assignment for the link $l_{m,n}$,
and $\bm{\varphi}$ to refer to per-topic link distribution.
The generative process of the Link LDA is described as follows:
\begin{enumerate}[1)]
    \item For each topic $k\in \left [ 1,K \right ]$
    \begin{enumerate}
        \item Draw topic-word distribution $\bm{\phi}_{k}\sim Dirichlet\left ( \beta  \right )$
        \item Draw topic-link distribution $\bm{\varphi}_{k}\sim Dirichlet\left ( \gamma  \right )$
    \end{enumerate}
    \item For each document $m \in \left [ 1,M \right ]$
    \begin{enumerate}
        \item Draw $\bm{\theta} _{m} \sim Dirichlet\left ( \alpha   \right )$
        \item For each word  $w_{m,n}$ in document $m$
        \begin{enumerate}
            \item Draw the topic $z_{m,n}\sim Multinomial(\bm{\theta}_{m})$
            \item Draw the word $w_{m,n}\sim Multinomial(\bm{\phi}_{z_{m,n}}) $
        \end{enumerate}
        \item For each link  $l_{m,n}$ for document $m$
        \begin{enumerate}
            \item Draw the topic $x_{m,n}\sim Multinomial(\bm{\theta}_{m})$
            \item Draw the link $l_{m,n}\sim Multinomial(\bm{\varphi}_{x_{m,n}}) $
        \end{enumerate}
     \end{enumerate}
\end{enumerate}
Same as the orginal paper \citep{erosheva2004mixed}, in estimation, we utilize the collapsed Gibbs sampling where latent
variables $\bm{z}$ and $\bm{x}$ are sampled iteratively.
Specifically, we first sample the topic assignments $z_{m,n}$ for the word $w_{m,n}$ according to the posterior distribution:
\begin{equation} \label{eq:topic assignment for word Link}
p(z_{m,n}=k|w_{m,n}=v,\bm{z}_{-m,n},\bm{w}_{-m,n},\bm{x},\alpha,\beta ) \propto \frac{n_{k,\left ( -m,n \right )}^{v}+\beta }{n_{k,\left ( -m,n \right )}^{\left ( \ast  \right )}+V\beta }\left ( n_{m,\left ( -m,n \right )}^{k}+c_{m}^{k}+\alpha  \right )
\end{equation}
Next, we sample the topic assignments $x_{m,e}$ for the link on the basis of posterior distribution: $p(x_{m,e}=k|l_{m,e}=l,\bm{x}_{-m,e},\bm{l}_{-m,e},\bm{z},\alpha,\gamma  )$:
\begin{equation} \label{eq:topic assignment for link Link}
p(x_{m,e}=k|l_{m,e}=l,\bm{x}_{-m,e},\bm{l}_{-m,e},\bm{z},\alpha,\gamma  ) \propto \frac{c_{k,\left ( -m,e \right )}^{l}+\gamma  }{c_{k,\left ( -m,e \right )}^{\left ( \ast  \right )}+L\gamma  }\left ( c_{m,\left ( -m,e \right )}^{k}+n_{m}^{k}+\alpha  \right )
\end{equation}
where $n_k^v$ is the number of times word $v$ assigned to topic $k$. $n_k^{( \ast ) }$ denotes the total number of words assigned to topic $k$. $n_m^k$ is the number of words in document $m$ assigned to topic $k$. $c_m^k$ is the number of links for document $m$ assigned to topic $k$. $c_k^l$ is the number of times link $l$ assigned to topic $k$. $c_k^{(\ast )}$ is the total number of links assigned to topic $k$. The notation $-m,n$ refers to the corresponding count excluding the current assignment of the word $w_{m,n}$. The notation $-m,e$ refers to the corresponding count excluding the current assignment of the link $l_{m,e}$.

Algorithm 10 summarizes the procedure of collapsed Gibbs sampling for Link LDA. After the sampling algorithm has achieved convergece
with appropriate iterations,
we can estimate the parameters $\bm{\theta}$, $\bm{\phi}$ and $\bm{\varphi}$ via the following equations:
\begin{equation} \label{eq:theta for Link}
\widehat{\theta}_{m,k}=\frac{n_{m}^{k}+c_{m}^{k}+\alpha }{\sum_{k^{'}=1}^{K}\left ( n_{m}^{k^{'}}+c_{m}^{k^{'}} \right )+K\alpha }
\end{equation}
\begin{equation} \label{eq:phi for Link}
\widehat{\phi } _{k,v}= \frac{n_{k}^{v} + \beta  }{n_{k}^{ \left ( \ast  \right )}+V\beta }
\end{equation}
\begin{equation} \label{eq:varphi for Link}
\widehat{\varphi } _{k,l}= \frac{c_{k}^{l} + \gamma }{c_{k}^{ \left ( \ast  \right )}+L\gamma }
\end{equation}
\begin{table}
\centering
\begin{tabular}{lllp{7.4cm}}
\hline
{\textbf{Algorithm 10:}} Procedure of collapsed Gibbs sampling for link LDA \\ \hline
{\textbf{Input:}} topic number $K$, hyperparameters $\alpha$, $\beta$ and $\gamma$, number of iterations $N_{iter}$ \\
{\textbf{Output:}} $\bm{\theta}$, $\bm{\phi}$ and $\bm{\varphi}$ \\
1. Randomly initialize the topic assignment $z_{m,n}$ for each word and the topic assignment $x_{m,e}$  \\
\qquad for each link; update the count $n_m^k$, $c_m^k$, $n_k^v$, $n_k^{(\ast) }$, $c_k^l$, $c_k^{(\ast)}$ \\
2. For $iter=1$ to $N_{iter}$ \\
\qquad   -For each document $m \in [1,M]$ \\
\qquad \qquad -For each word  $w_{m,n}$ in document $m$  \\
\qquad  \qquad \qquad Sample the topic $k$ using Equation (\ref{eq:topic assignment for word Link}) \\
\qquad  \qquad \qquad Update the count $n_m^k$, $n_k^v$, $n_k^{(\ast) }$   \\
\qquad \qquad -For each link  $l_{m,e}$ for document $m$  \\
\qquad  \qquad \qquad Sample the topic $k$ using Equation (\ref{eq:topic assignment for link Link}) \\
\qquad  \qquad \qquad Update the count $c_m^k$, $c_k^l$, $c_k^{(\ast)}$  \\
3. Estimate $\bm{\theta}$ using Equation (\ref{eq:theta for Link}), $\bm{\phi}$ using Equation (\ref{eq:phi for Link}) and $\bm{\varphi}$ using Equation (\ref{eq:varphi for Link}) \\
  \hline
\end{tabular}
\end{table}

\subsection{Labeled LDA}
To incorporate supervision information, Labeled LDA \citep{ramage2009labeled} is proposed as supervised topic model for documents
 with multiple labels. It learns word-label correspondences directly by introducing a one-to-one correspondence between latent topics
 and document labels. Besides, the topics of each document are restricted to be defined only over the document label sets. The extension
 allows the Labeled LDA gain advantages
over standard LDA such as the ability to assign a document's words to its labels as well as enhanced interpretability of topics.

For Labeled LDA, let $\bm{\Lambda} _{m}=\left \{ l_{1},l_2,\cdots ,l_K \right \}$ denote a list of binary topic absence indicators for document $m$, where
each $l_{k}\in \left \{ 0,1 \right \}$. The number of topics is equal to the number of unique labels $K$ in the corpus. $\Psi _{k}$ is the label prior for topic $k$. $\bm{\lambda} _{m}=\left \{ k|\Lambda _{m}^{k}=1 \right \}$ denotes the vector of labels of document $m$, where $\Lambda _{m}^{k} \in \bm{\Lambda} _{m}$. $\bm{L}^{(m) }$ represents a document-specific label projection matrix with the size of $P_m\times K$, where $P_{m}=\left | \bm{\lambda}^{\left ( m \right )} \right |$. For each row $i \in\left [ 1,P_m \right ]$ and column $k \in \left [ 1,K \right ]$, $\bm{L}^{(m)}$ can be defined as follows:
\begin{equation} \label{eq:L}
L_{ik}^{\left ( m \right )}=
\left\{
\begin{aligned}
1 \: \: \: \: \: \: \: \: \: \: \: \: \: \: \: if \,\, \lambda _{i}^{\left ( m \right )}=k \\
0 \: \: \: \: \: \: \: \: \: \: \: \: \: \: \: otherwise\\
\end{aligned}
\right.
\end{equation}
To restrict the topic of document $m$ to its own labels, Labeled LDA introduces a lower dimensional vector
$\bm{\alpha}^{\left ( m \right )}=\bm{L}^{\left ( m \right )}\times \bm{\alpha} $ where $\bm{\alpha} =\left ( \alpha_{1},\alpha _{2},\cdots ,\alpha _{K}  \right )$
is the Dirichlet topic prior. The generative process of the Labeled LDA is described as follows:
\begin{enumerate}[1)]
    \item For each topic $k\in \left [ 1,K \right ]$
    \begin{enumerate}
        \item Draw topic-word distribution $\bm{\phi}_{k}\sim Dirichlet\left ( \beta  \right )$
    \end{enumerate}
    \item For each document $m \in \left [ 1,M \right ]$
    \begin{enumerate}
        \item For each topic $k\in \left [ 1,K \right ]$
        \begin{enumerate}
            \item Draw $\Lambda _{m}^{k}\in\left \{ 0,1 \right \}\sim Bernoulli\left ( \Psi _{k} \right )$
        \end{enumerate}
        \item  Generate $\bm{L}^{(m) }$ and $\bm{\alpha}^{\left ( m \right )}$
         \item Draw $\bm{\theta} _{m} \sim Dirichlet\left ( \bm{\alpha}^{\left ( m \right )}   \right )$
        \item For each word  $w_{m,n}$ in document $m$
        \begin{enumerate}
            \item Draw the topic $z_{m,n}\sim Multinomial(\bm{\theta}_{m})$
            \item Draw the word $w_{m,n}\sim Multinomial(\bm{\phi}_{z_{m,n}}) $
        \end{enumerate}
     \end{enumerate}
\end{enumerate}
From the generative process, we find that Labeled LDA is same as traditional LDA, excluding the constrain of the topic prior $\bm{\alpha}^{(m) }$. Same as \cite{ramage2009labeled}, we use collapsed Gibbs sampling for topic assignments. The probability for sampling topic assignment $z_{m,n}$
is given by:
\begin{equation} \label{eq:topic assignment Labeled LDA}
p\left ( z_{m,n}=k|w_{m,n}=v,\bm{w}_{-m,n},\bm{z}_{-m,n},\beta ,\bm{\alpha} ^{\left ( m \right )} \right )\propto \frac{n_{k,\left ( -m,n \right )}^{v}+\beta }{n_{k,\left ( -m,n \right )}^{\left ( \ast  \right )}+V\beta }\frac{n_{m,\left ( -m,n \right )}^{k}+\alpha _{k}}{\sum_{k^{'}=1}^{K}\left ( n_{m,\left ( -m,n \right )}^{k^{'}} +\alpha _{k^{'}}   \right )}
\end{equation}
Algorithm 11 summarizes the procedure of collapsed Gibbs sampling for Labeled LDA. We can estimate the parameters $\bm{\theta}$, and $\bm{\phi}$  via the following equations:
\begin{equation} \label{eq:theta for Labeled LDA}
\widehat{\theta}_{m,k}=\frac{n_{m}^{k}+\alpha _{k}}{\sum_{k^{'}=1}^{K}\left ( n_{m}^{k^{'}} +\alpha _{k^{'}}   \right )}
\end{equation}
\begin{equation} \label{eq:phi for Labeled LDA}
\widehat{\phi } _{k,v}= \frac{n_{k}^{v} + \beta  }{n_{k}^{ \left ( \ast  \right )}+V\beta }
\end{equation}
\begin{table}
\centering
\begin{tabular}{lllp{7.4cm}}
\hline
{\textbf{Algorithm 11:}} Procedure of collapsed Gibbs sampling for Labeled LDA \\ \hline
{\textbf{Input:}} hyperparameters $\alpha$, and $\beta$, number of iterations $N_{iter}$ \\
{\textbf{Output:}} $\bm{\theta}$, $\bm{\phi}$ \\
1. 	Scan the corpus and determine the number of unique labels $K$ \\
2. Randomly initialize the topic assignment $z_{m,n}$ for each word according to labels of document $m$,  \\
\qquad and update the count $n_m^k$, $n_k^v$, $n_k^{(\ast) }$ \\
2. For $iter=1$ to $N_{iter}$ \\
\qquad   -For each document $m \in [1,M]$ \\
\qquad \qquad -For each word  $w_{m,n}$ in document $m$  \\
\qquad  \qquad \qquad Sample the topic $k$ using Equation (\ref{eq:topic assignment Labeled LDA}) \\
\qquad  \qquad \qquad Update the count $n_m^k$, $n_k^v$, $n_k^{(\ast) }$   \\
3. Estimate $\bm{\theta}$ using Equation (\ref{eq:theta for Labeled LDA}) and $\bm{\phi}$ using Equation (\ref{eq:phi for Labeled LDA}) \\
  \hline
\end{tabular}
\end{table}
\subsection{PLDA}
One of the underlying assumption of Labeled LDA is that there is no sub-topic within a label and all topics are covered by the labels
 occuring in the documents. This rigid assumption can be inappropriate and cause compromized results.  \cite{ramage2011partially} propose
 the Partially Labeled Dirichlet Allocation (PLDA), which extends Labeled LDA to incorporate per-label latent topics where each label contains
 multiple topics but each topic belongs to only one label. Also, PLDA gives a background label with a set of background topics.

For PLDA, let $\bm{L}$ denote a set of labels (indexed by $1,2,\cdots,L$). The index of $L$ represents the background label. Each label $l \in L$ corresponds to a set of topics $K_l$  (indexed by $1,2,\cdots,K_l$). $\bm{\Lambda} _{m}$ is a sparse binary vector that represents a subset of label classes of document $m$. $\bm{\phi} _{l,k}$ denotes the label-topic-word distribution. $\bm{\varphi} _{m}$ is a document-level mixture of observed labels that is sampled from a multinomial of size $\left | \bm{\Lambda} _{m} \right |$ from a Dirichlet prior $\bm{\alpha} ^{\left ( L \right )}$. And each $\varphi _{m,l}\in \bm{\varphi} _{m}$  represents the document's probability of using label $l \in \bm{\Lambda} _{m}$. $\bm{\theta} _{m,l}$ is a document-level mixture over topics $\bm{K}_l$, and it is sampled from a symmetric Dirichlet prior $\alpha$. The generative process of the PLDA is described as follows:
\begin{enumerate}[1)]
    \item For each label $l\in \left [ 1,L \right ]$
    \begin{enumerate}
        \item For each topic $k\in \left [ 1,K_l \right ]$
        \begin{enumerate}
              \item Draw label-topic-word distribution $\bm{\phi}_{l,k}\sim Dirichlet\left ( \beta  \right )$
         \end{enumerate}
    \end{enumerate}
    \item For each document $m \in \left [ 1,M \right ]$
    \begin{enumerate}
        \item For each label $l\in \left [ 1,L \right ]$
        \begin{enumerate}
            \item Draw $\Lambda _{m}^{l}\in\left \{ 0,1 \right \}\sim Bernoulli\left ( \gamma  \right )$
            \item Draw $\bm{\theta} _{m,l} \sim Dirichlet\left ( \alpha   \right )$
        \end{enumerate}
        \item  Draw $\bm{\varphi} _{m} \sim Dirichlet\left ( \bm{\alpha}^{(L)}   \right )$
        \item For each word  $w_{m,n}$ in document $m$
        \begin{enumerate}
            \item Draw the label $l_{m,n}\sim Multinomial(\bm{\varphi} _{m})$
            \item Draw the topic $z_{m,n}\sim Multinomial(\bm{\theta}_{m,l_{m,n}})$
            \item Draw the word $w_{m,n}\sim Multinomial(\bm{\phi}_{l_{m,n},z_{m,n}}) $
        \end{enumerate}
     \end{enumerate}
\end{enumerate}
Same as \cite{ramage2011partially}, we use collapsed Gibbs sampling to track the label and topic assignments.
Since each label has its own distinct subset of topics, the topic assignment is adequate to ascertain which label is assigned to different
words. Hence, in implementation, we just allocate memory to record the topic assignments. Using conjugate priors and Bayes rule,
 we can obtain the conditional probability for sampling the label assignment $l_{m,n}$ and topic assignment $z_{m,n}$:
\begin{equation} \label{eq:topic assignment for PLDA}
\begin{split}
p\left ( l_{m,n}=l, z_{m,n}=k|w_{m,n} =v,\bm{w}_{-m,n},\bm{l}_{-m,n},\bm{z}_{-m,n},\alpha ,\beta  \right ) \propto & I\left [ l \in\Lambda _{m}\cap k \in K_{l} \right ] \\
& \left ( n_{m,\left ( -m,n \right )}^{\left ( l,k \right )}+\alpha  \right )\frac{n_{m,(-m,n)}^{v}+\beta }{n_{l,k,(-m,n)}^{\left ( \ast  \right )}+V\beta }
\end{split}
\end{equation}
where $I[\ast ]$ is the indicator function. $n_{m}^{\left ( l,k \right )}$ is the number of word tokens in document $m$ assigned to topic $k \in \bm{K}_{l}$. $n_{l,k}^v$ is the number of tokens of word $v$ assigned to topic $k \in \bm{K}_{l}$. $n_{l,k}^{(*) }$ is the total number of words assigned to topic $k \in \bm{K}_{l}$. The notation $-m,n$ refers to the corresponding count excluding the current assignment of the word $w_{m,n}$.

Since each topic is related to only a single label, we can use empirical estimation to obtain document-topic proportion $\widehat{\theta}_{m,k}$ and topic-word proportion $\widehat{\phi } _{k,v}$.
\begin{equation} \label{eq:theta for PLDA}
\widehat{\theta}_{m,k}=\frac{n_{m}^{k}+\alpha _{k}}{\sum_{k^{'}=1}^{K}\left ( n_{m}^{k^{'}} +\alpha _{k^{'}}   \right )}
\end{equation}
\begin{equation} \label{eq:phi for PLDA}
\widehat{\phi } _{k,v}= \frac{n_{k}^{v} + \beta  }{n_{k}^{ \left ( \ast  \right )}+V\beta }
\end{equation}
where $K=\sum_{l=1}^{L}K_l$ is the total number of topics associated with labels. Algorithm 12 presents the procedure of collapsed Gibbs sampling for PLDA.
\begin{table}
\centering
\begin{tabular}{lllp{7.4cm}}
\hline
{\textbf{Algorithm 12:}} Procedure of collapsed Gibbs sampling for PLDA \\ \hline
{\textbf{Input:}} hyperparameters $\alpha$ and $\beta$,the topic number of $K_{l}$ each label, number of iterations $N_{iter}$ \\
{\textbf{Output:}} $\bm{\theta}$, $\bm{\phi}$ \\
1. 	Scan the corpus and determine the number of unique labels $\bm{L}$ \\
2. 	Allocate memory to represent each label $l$ and its topics $\bm{K}_l$ \\
3. Randomly initialize the topic assignment $z_{m,n}$ for each word according to labels of document $m$,  \\
\qquad and update the count $n_m^{(l,k,)}$, $n_{l,k}^v$, $n_{l,k}^{(\ast) }$ \\
4. For $iter=1$ to $N_{iter}$ \\
\qquad   -For each document $m \in [1,M]$ \\
\qquad \qquad -For each word  $w_{m,n}$ in document $m$  \\
\qquad  \qquad \qquad Sample the topic $k$ using Equation (\ref{eq:topic assignment for PLDA}) \\
\qquad  \qquad \qquad Update the count $n_m^{(l,k,)}$, $n_{l,k}^v$, $n_{l,k}^{(\ast) }$   \\
5. Estimate $\bm{\theta}$ using Equation (\ref{eq:theta for PLDA}) and $\bm{\phi}$ using Equation (\ref{eq:phi for PLDA}) \\
  \hline
\end{tabular}
\end{table}

\subsection{Dual-Sparse Topic Model}
The Dual-Sparse Topic Model is an effective approach for short documents modeling, which is proposed by \cite{lin2014dual}. Classical topic models such as LDA usually utilize Dirichlet prior to smooth the document-topic distributions and the topic-word distributions, thus fail to control the posterior sparsity of the topic mixtures and the word distributions. In reality, a short document generally focuses on a small number of topics, and a topic is often characterized by a narrow range of words in the vocabulary. Thus, the topic mixtures and the word distributions should have the feature of skewness or sparsity. In order to induce this sparsity, Dual-Sparse Topic Model introduces a "Spike and Slab" prior to decouple the sparsity and smoothness of the document-topic distributions and topic-word distributions. This model is easy enough to learn both a few focused topics for a document and focused words for a topic.

For Dual-Sparse Topic Model, let $\alpha_{mk}$ denote a topic selector that uses a binary variable to indicate whether a topic $k$ is associated with document $m$. $\alpha_{mk}$ is drawn from a Bernoulli distribution with a Bernoulli parameter $a_m$. $\beta_{kv}$ is a word selector that indicates whether the word $v$ is involved in topic $k$. $\beta_{kv}$ is drawn from a Bernoulli distribution with a Bernoulli parameter $b_k$. $\pi$ and $\bar{\pi}$ are the topic smoothing prior and the weak topic smoothing prior, respectively. $\gamma $ and $\bar{\gamma }$ are the word smoothing prior and the weak word smoothing prior, respectively. Since $\bar{\pi}<<\pi$ and $\bar{\gamma }<<\bar{\gamma }$, the model can easily induce the sparsity while also avoiding the ill-definition of the distributions. We describe the generative process of the Dual-Sparse Topic Model as follows:
\begin{enumerate}[1)]
    \item For each topic $k\in \left [ 1,K \right ]$
    \begin{enumerate}
        \item For each topic $k\in \left [ 1,K_l \right ]$
        \begin{enumerate}
              \item Draw $b_{k}\sim Beta\left ( x,y \right )$
              \item For each word $v \in \left [ 1,V \right ]$
              \begin{enumerate}
                    \item Draw the word selector $\beta _{kv}\sim Bernoulli(b_k )$, $\bm{\beta} _{k}=\left \{ \beta _{kv}=1 \right \}_{v=1}^{V}$
                    \item Determine the set of focused words: $B_{k}=\left \{ v:\beta _{kv}=1 \right \}$
                    \item Draw topic-word distribution $\bm{\phi}_{k}\sim Dirichlet\left (\gamma \bm{\beta} _{k}+ \bar{\gamma }\bm{1} \right )$
              \end{enumerate}
         \end{enumerate}
    \end{enumerate}
    \item For each document $m \in \left [ 1,M \right ]$
    \begin{enumerate}
        \item Draw $a_m \sim Beta(s,t)$
        \item For each topic $k\in \left [ 1,K \right ]$
        \begin{enumerate}
            \item Draw the topic selector $\alpha _{mk}\sim Bernoulli\left ( a_m \right )$, $\bm{\alpha} _{m}=\left \{ \alpha _{mk} \right \}_{k=1}^{K}$
             \item Determine the set of focused topics: $A_{m}=\left \{ k:\alpha _{mk}=1 \right \}$
        \end{enumerate}
        \item Draw $\bm{\theta} _{m} \sim Dirichlet\left ( \pi \bm{\alpha} _{m}+ \bar{\pi} \bm{1} \right )$
        \item For each word  $w_{m,n}$ in document $m$
        \begin{enumerate}
            \item Draw the topic $z_{m,n}\sim Multinomial(\bm{\theta}_{m})$
            \item Draw the word $w_{m,n}\sim Multinomial(\bm{\phi}_{z_{m,n}}) $
        \end{enumerate}
     \end{enumerate}
\end{enumerate}
For large document collections, \cite{lin2014dual} use Zero-Order Collapsed Variational Bayesian Inference Algorithm (CVB0) for the posterior inference. The parameters that need to infer contain topic selector $\bm{\alpha}$,  word selector $\bm{\beta}$ and topic assignment $\bm{z}$.

\textbf{Variational Bernoulli distribution for} $\bm{\alpha}$:
\begin{equation} \label{eq:Bernoulli distribution 1}
\begin{split}
  \hat{\alpha}_{mk} &=\frac{\tilde{\alpha}_{mk}^{1} }{\tilde{\alpha}_{mk}^{1}+\tilde{\alpha}_{mk}^{0}} \\
  \tilde{\alpha}_{mk}^{1} & =(s+\hat{A}_{m}^{(-mk)})\Gamma (\hat{n}_{m}^{k}+\pi +\bar{\pi} )B(\pi+K\bar{\pi}+\pi\hat{A}_{m}^{(-mk)},\hat{n}_{m}^{(\ast) }+\pi\hat{A}_{m}^{(-mk)}+K\bar{\pi}) \\
 \tilde{\alpha}_{mk}^{0} & =(t+K-1-\hat{A}_{m}^{(-mk)})\Gamma (\pi +\bar{\pi} )B(K\bar{\pi}+\pi\hat{A}_{m}^{(-mk)},n_{m}^{(\ast) }+\pi+\pi \hat{A}_{m}^{(-mk)}+K \bar{\pi} )
\end{split}
\end{equation}
\textbf{Variational Bernoulli distribution for} $\bm{\beta}$:
\begin{equation} \label{eq:Bernoulli distribution 2}
\begin{split}
  \hat{\beta }_{kv} &=\frac{\tilde{\beta }_{kv}^{1} }{\tilde{\beta }_{kv}^{1}+\tilde{\beta }_{kv}^{0}} \\
  \tilde{\beta }_{kv}^{1} & =(x+\hat{B}_{k}^{(-kv)})\Gamma (\hat{n}_{k}^{v}+\gamma  +\bar{\gamma } )B(\gamma +V\bar{\gamma }+\pi\hat{B}_{k}^{(-kv)},\hat{n}_{k}^{(\ast) }+\pi\hat{B}_{k}^{(-kv)}+V\bar{\gamma }) \\
 \tilde{\beta }_{kv}^{0} & =(y+V-1-\hat{B}_{k}^{(-kv)})\Gamma (\gamma  +\bar{\gamma } )B(V\bar{\gamma }+\gamma \hat{B}_{k}^{(-kv)},n_{k}^{(\ast) }+\gamma +\gamma  \hat{B}_{k}^{(-kv)}+V \bar{\gamma } )
\end{split}
\end{equation}
\textbf{Variational Multinomial distribution for} $\bm{z}$
\begin{equation} \label{eq:Multinomial distribution z}
\kappa _{m,n,k}=\hat{q}(z_{m,n}=k)\propto (\hat{n}_{m,(-m,n)}^{k}+\pi \hat{\alpha }_{mk}+\bar{\pi})\frac{\hat{n}_{k,(-m,n)}^{w_{m,n}}+\gamma \hat{\beta}_{kv}+ \bar{\gamma }  }{\hat{n}_{k,(-m,n)}^{(\ast )}+\gamma \hat{B}_{k}+ V\bar{\gamma }}
\end{equation}
where $\hat{\alpha}_{mk}$, $ \hat{\beta }_{kv}$ and $\kappa _{m,n,k}$ are variational parameters for topic selector $\alpha_{m,k}$,
word selector $\beta_{kv}$ and topic assignment $z_{m,n}$, respectively. $\Gamma(\cdot )$ and $\mathrm{B}(\cdot )$ denote Gamma function and
Beta function, respectively. $\hat{A}_{m}=\sum_{{k}'=1}^{K}\hat{\alpha }_{m{k}'}$ indicates the expected number of focused topics for document
$m$. $\hat{n}_{m}^{k}=\sum_{{n}'}\kappa _{m,{n}',k}$ denotes the expected number of word tokens assigned to topic $k$ in document $m$.
$\hat{n}_{m}^{(\ast )}=\sum_{{n}',{k}'}\kappa _{m,{n}',{k}'}$ denotes the expected number of word tokens in document $m$. $\hat{B}_{m}=\sum_{{v}'=1}^{V}\hat{\beta }_{kv}$  indicates the expected number of focused words for topic $k$. $\hat{n}_{k}^{v}=\sum_{{m}',{n}'}I\left [ w_{{m}',{n}'}=v \right ]\kappa _{{m}',{n}',k}$ is the expected number of tokens of word $v$ assigned to topic $k$. $\hat{n}_{k}^{(\ast )}=\sum_{{m}',{n}'}\kappa _{{m}',{n}',k}$ is the expected number of words assigned to topic $k$. $V$ denotes the vocabulary size.
The notation $-mk$, $-kv$, and $-m,n$ mean $\alpha _{mk}$, $\beta  _{kv}$ and $w_{m,n}$ are excluded, respectively.

When the algorithm has reached convergence after an appropriate number of
iterations, we can estimate the parameters $\bm{\theta}$ and $\bm{\phi}$ via the following equations:
\begin{equation} \label{eq:theta for DSTM}
\widehat{\theta}_{m,k}=\frac{\hat{n}_{m}^{k}+\pi \hat{\alpha}_{mk}+\bar{\pi }}{\hat{n}_{m}^{\left ( \ast  \right )}+\pi \hat{A}_{m}+K\bar{\pi }}
\end{equation}
\begin{equation} \label{eq:phi for DSTM}
\widehat{\phi } _{k,v}= \frac{\hat{n}_{k}^{v}+\gamma  \hat{\beta }_{kv}+\bar{\gamma  }}{\hat{n}_{k}^{\left ( \ast  \right )}+\pi\gamma \hat{B}_{k}+V\bar{\gamma }}
\end{equation}
According to the estimated parameters $\bm{\alpha}$ and $\bm{\beta}$, we can measure the sparsity ratio of a document-topic distribution and a topic-word distribution as follows:
\begin{equation} \label{eq:sparsity for document-topic distribution}
sparsityD(m)=1-\frac{\hat{A}_{m}}{K}
\end{equation}
\begin{equation} \label{eq:sparsity for topic-word distribution}
sparsityT(k)=1-\frac{\hat{B}_{k}}{V}
\end{equation}
The average sparsity ratio of all documents can be calculated as follows:
\begin{equation} \label{eq:average sparsity for document-topic distribution}
Asparsity-document=\frac{1}{M}\sum_{m=1}^{M}sparsity(m)
\end{equation}
\begin{equation} \label{eq:average sparsity for topic-word distribution}
Asparsity-topic=\frac{1}{K} \sum_{k=1}^{K}sparsity(k)
\end{equation}
Algorithm 13 shows the procedure of CVB0 for Dual-Sparse Topic Model.
\begin{table}
\centering
\begin{tabular}{lllp{7.4cm}}
\hline
{\textbf{Algorithm 13:}} Procedure of CVB0 for Dual-Sparse Topic Model \\ \hline
{\textbf{Input:}}  hyperparameters $x$, $y$, $\gamma $, $\bar{\gamma }$, $s$, $t$, $\pi $ and $\bar{\pi }$, $K$, number of iterations $N_{iter}$ \\
{\textbf{Output:}} $\bm{\theta}$, $\bm{\phi}$, $sparsity(m)$, $sparsity(k)$, $Asparsity-document$, $Asparsity-topic$ \\
1. Randomly initialize the variational parameter $\kappa_{m,n,k}$, \\
\qquad and update the value of $\hat{n}_{m}^{k}$, $\hat{n}_{ m}^{\left ( \ast \right )}$, $\hat{n}_{k}^{v}$ and $\hat{n}_{ k}^{\left ( \ast \right )}$ \\
2. 	Randomly initialize $\hat{\alpha} _{mk}$, $\hat{\beta} _{kv}$ and update the value $\hat{A}_{m}$ and $\hat{B}_{k}$ \\
3. For $iter=1$ to $N_{iter}$ \\
\qquad   -For each document $m \in [1,M]$ \\
\qquad \qquad - For each topic $k \in [1,K]$   \\
\qquad  \qquad \qquad Update $\hat{\alpha} _{mk}$ using Equation (\ref{eq:Bernoulli distribution 1}) \\
\qquad  \qquad \qquad Update $\hat{A}_{m}$ \\
\qquad  -For each topic $k \in [1,K]$  \\
\qquad  \qquad  -For each word $v$ in vocabulary    \\
\qquad  \qquad  \qquad Update $\hat{\beta} _{kv}$ using Equation (\ref{eq:Bernoulli distribution 2})\\
\qquad   -For each document $m \in [1,M]$ \\
\qquad \qquad  -For each word  $w_{m,n}$ in document $m$ \\
\qquad \qquad \qquad -For each topic $k \in [1,K]$ \\
\qquad \qquad \qquad \qquad Update $\hat{\alpha} _{mk}$, $\hat{\beta} _{kv}$ using Equation Equation (\ref{eq:Multinomial distribution z})\\
4. Estimate $\bm{\theta}$ using Equation (\ref{eq:theta for DSTM}), $\bm{\phi}$ using Equation (\ref{eq:phi for DSTM}), $sparsity(m)$ using Equation  (\ref{eq:sparsity for document-topic distribution}), \\
\qquad $sparsity(k)$ using Equation (\ref{eq:sparsity for topic-word distribution}), $Asparsity-document$ using Equation (\ref{eq:average sparsity for document-topic distribution}), and \\ \qquad $Asparsity-topic$ using Equation (\ref{eq:average sparsity for topic-word distribution})\\
  \hline
\end{tabular}
\end{table}

\section{The Use of TopicModel4J} \label{sec:use}
\subsection{Download Package}
The \pkg{TopicModel4J} is implemented by the \proglang{Java} language. And it has been released under the GNU General Public License (GPL). All the source code for this package can be downloaded from \url{https://github.com/soberqian/TopicModel4J}. The researchers can run this package on any platforms (e.g. Linux and Windows) that support \proglang{Java} 1.8. The dependencies for this package contain \pkg{commons-math3} and \pkg{Stanford CoreNLP}\citep{manning2014stanford}.

\subsection{Data Pre-processing} \label{SenLDA}
For the tasks of NLP and text mining applications, the data pre-preprocessing is an important procedure before applying various machine learning models. It make the input data more amenable for analysis and reliable discoveries. In \pkg{TopicModel4J}, we provide an easy-to-use interface for common text preprocessing techniques such as lowercasing, lemmatization and stopword removal and noise removal. Next, we give two examples for the use of data preprocessing in \pkg{TopicModel4J}. In the first example, we deal with a single document by the following code:
\begin{Code}
String line = "http://t.cn/RAPgR4n Artificial intelligence is a known phenomenons "
			+ "in the world today. Its root started to build years";
//get all words for this document
ArrayList<String> words = new ArrayList<String>();
//lemmatization
FileUtil.getlema(line, words);
//remove noise words using the default stop words
String text = FileUtil.RemoveNoiseWord(words);
//remove noise words using the customized stop words
String text_c = FileUtil.RemoveNoiseWord(words,"data/stopwords");
//print results
System.out.println(text);
System.out.println(text_c);
\end{Code}

By the above code, we can remove uninformative words from the original document using the default stop words defined by \pkg{TopicModel4J}. The
default stop words contain 524 common words (e.g., "a" and "the") that are given by \pkg{BOW} library (\url{http://www.cs.cmu.edu/~mccallum/bow/}).
In order to meet the particular needs of researchers, \pkg{TopicModel4J} also provides an option to define a customized set of stop word list.
 For example, the  \code{"data/stopwords"} in the above code is a file that includes many user-defined stop words (e.g. "year" and "today" ). Results after data preprocessing are shown below:
\begin{Code}
artificial intelligence phenomenon world today root start build year
artificial intelligence phenomenon world root start build
\end{Code}

The second example is to show how mutiple documents are processed. In \pkg{TopicModel4J}, documents are merged in a single file where each line represent one document accordingly.
For example, the file \code{"rawdata"} consists of the following 4 documents:
\begin{Code}
Your review helps others learn about great local businesses.
I am a new ""member"" and let me tell you
I like the store in general. But the people who attend the Dim Sum Section are horrible.
they have a tough competition compared to all the other restaurants in the valley.
\end{Code}

Here we give the code to deal with this file.
\begin{Code}
//read raw data from a file
ArrayList<String> docLines = new ArrayList<String>();
FileUtil.readLines("data/rawdata", docLines, "gbk");
ArrayList<String> doclinesAfter = new ArrayList<String>();
for(String line : docLines){
    //get all word for a document
    ArrayList<String> words = new ArrayList<String>();
    //lemmatization
    FileUtil.getlema(line, words);
    //remove noise words
    String text = FileUtil.RemoveNoiseWord(words);
    doclinesAfter.add(text);
}
// write the processed data to a new file
FileUtil.writeLines("data/rawdataprocess", doclinesAfter, "gbk");
\end{Code}

\begin{figure}[t!]
\centering
\includegraphics{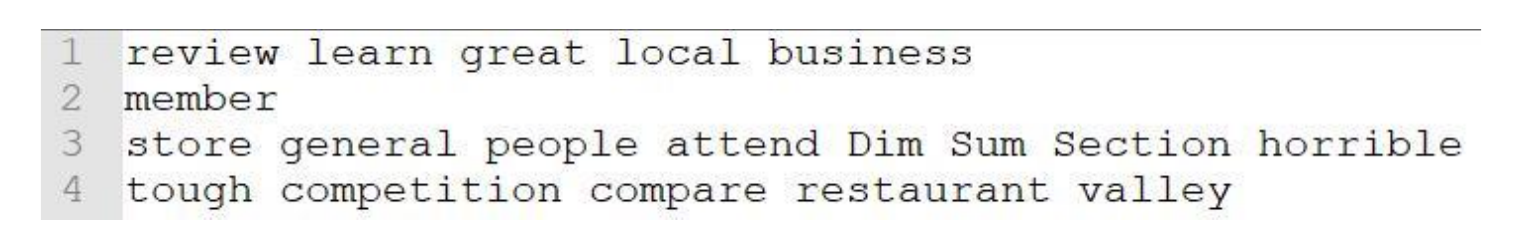}
\caption{\label{fig:fileprocessing} The content of the file "rawdataprocess"}
\end{figure}

Running this code, the processed data is saved to a new file \code{"rawdataprocess"}. And Figure~\ref{fig:fileprocessing} displays the results.

\subsection{Usage Examples}
\subsubsection{Data Set}
We provide 3 kind of data to construct the input data of these models in \pkg{TopicModel4J}.
\begin{itemize}
    \item Citation data. It consists of academic papers obtained from the ACM Digital Library (\url{https://dl.acm.org/}).
    Each paper has title, authors, abstracts, journal, year, and citation relations. This data is publicly available and can be downloaded from AMiner (\url{https://www.aminer.cn/citation}).
       \item Amazon review data. This dataset is a series of consumer reviews from Amazon ((\url{http://www.amazon.com})). It is a public data set and can be downloaded from \url{https://nijianmo.github.io/amazon/index.html}.
        In \pkg{TopicModel4J}, we collect the reviews in four categories: "Amazon Fashion", "Beauty", "Gift Cards" and "Software".
    \item ProgrammableWeb data. This dataset is crawled from the website ProgrammableWeb (\url{https://www.programmableweb.com/}). It contains the news of APIs (Application Programming Interfaces)
    and  labels for each news.

\end{itemize}
\subsubsection{Application of LDA}
In \pkg{TopicModel4J}, we implement a constructor method \code{GibbsSamplingLDA()} for collapsed Gibbs sampling:
\begin{Code}
public GibbsSamplingLDA(String inputFile, String inputFileCode, int topicNumber,
			double inputAlpha, double inputBeta, int inputIterations,
                               int inTopWords, String outputFileDir)
\end{Code}
where \code{inputFile} refers to the input composed of documents that have been processed.
The \code{inputFileCode} specifies the encoding format (e.g., utf-8 and gbk) of the input file.
The \code{topicNumber}, \code{inputAlpha}, \code{inputBeta} and \code{inputIterations} are the input parameters $K$, $\alpha$
, $\beta$ and $N_{iter}$ of this algorithm (See Algorithm 1).
The \code{inTopWords} determines the number of top words that are shown for each topic. The \code{outputFileDir} is the output directory.

\begin{figure}[t!]
\centering
\includegraphics{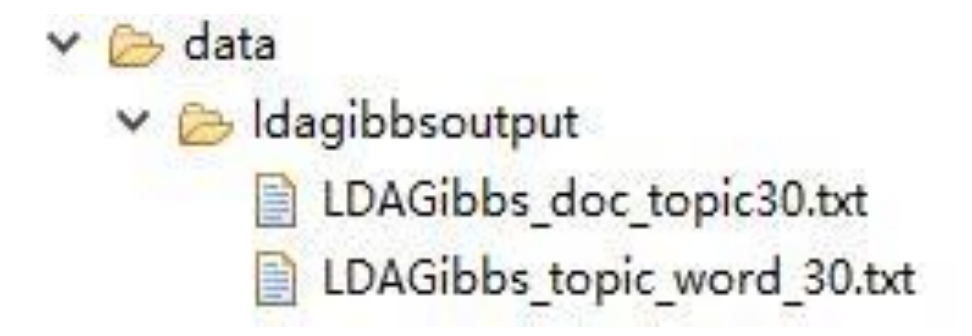}
\caption{\label{fig:ldagibbs} The output results of the collapsed Gibbs sampling for LDA}
\end{figure}

Using this constructor, we can construct an instance and call the LDA algorithm for processing text.
The following code shows how to use \code{GibbsSamplingLDA()} to discover topics from the abstracts of the citation data.
\begin{Code}
GibbsSamplingLDA lda = new GibbsSamplingLDA("data/rawdataProcessAbstracts",
"gbk", 30, 0.1, 0.01, 1000, 5, "data/ldagibbsoutput/");
lda.MCMCSampling();
\end{Code}
The input file \code{rawdataProcessAbstracts} includes 2,718 document abstracts. We give the format of the input file as follows:
\begin{Code}
paper present indoor navigation range strategy monocular ...
globalisation education increasingly topic discussion university ...
professional software development team base recognise sociotechnical ...
.
.
.
concept education education program lead graduate ready ...
\end{Code}
We obtain the results after running the code for \code{1000} iterations.
The results are saved to the output directory \code{"data/ldagibbsoutput/"} that contains two files (See Figure~\ref{fig:ldagibbs} ).

The file "LDAGibbs\_topic\_word\_30.txt" gives the five most frequent words (with probability) for each topic:
\begin{Code}
Topic:1
distribution :0.03822039986269391
model :0.034563622615869705
estimate :0.02783987090396713
estimation :0.024183093657142923
method :0.017223420832542014

Topic:2
service :0.040989429399418104
datum :0.032584529688293354
system :0.020034747927846815
user :0.019343934252959848
information :0.019228798640478686

Topic:3
algorithm :0.024406026950998437
time :0.023019373398992918
performance :0.015993662068831634
computational :0.012573249973884692
efficiency :0.011648814272547681
.
.
.
Topic:30
fuzzy :0.06345885891868693
method :0.029147466429561415
set :0.023490390965404524
decision :0.022014632148667942
criterion :0.020907813036115507
\end{Code}
The file "LDAGibbs\_doc\_topic30.txt" represents the topic distribution for each document:
\begin{Code}
Topic1 Topic2 Topic3 ... Topic30
0.0593 0.129 0.00116 ... 0.00116
0.00161 0.00161 0.291 ... 0.0177
.
.
.
0.002 0.142 0.0220 ... 0.00125
\end{Code}

\begin{figure}[t!]
\centering
\includegraphics{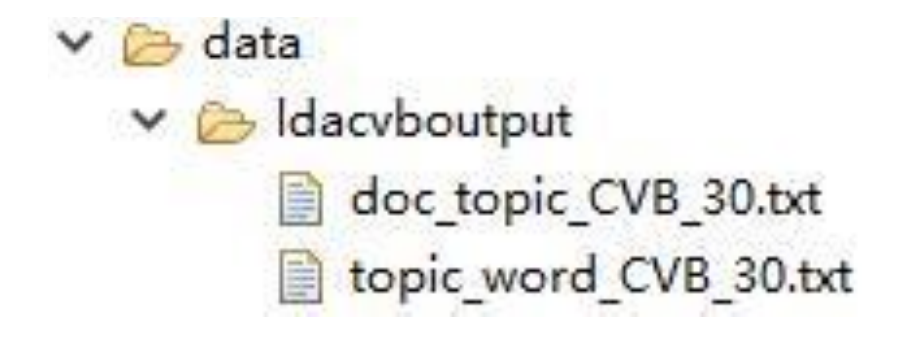}
\caption{\label{fig:ldacvb} The output results of the CVB for LDA}
\end{figure}

\pkg{TopicModel4J} also provides a constructor method \code{CVBLDA()} for the CVB inference:
\begin{Code}
public CVBLDA(String inputFile, String inputFileCode, int topicNumber,
              double inputAlpha, double inputBeta, int inputIterations,
                   int inTopWords,String outputFileDir)
\end{Code}
The meaning of parameters (i.e., \code{inputFile}, \code{inputFileCode}, etc.)
in \code{CVBLDA()} is the same as that in \code{GibbsSamplingLDA()}. The following code is to call the CVB inference for LDA:
\begin{Code}
CVBLDA cvblda = new CVBLDA("data/rawdataProcessAbstracts", "gbk", 30, 0.1,
				0.01, 1000, 5, "data/ldacvboutput/");
cvblda.CVBInference();
\end{Code}
We will obtain the results after \code{1000} iterations. The results are saved to the output
directory \code{"data/ldacvboutput/"} that contains two files (See Figure~\ref{fig:ldacvb} ).
Contents for the two files are similar to the output results of the collapsed Gibbs sampling for LDA.

\subsubsection{Application of Sentence-LDA}
Sentence-LDA has been widely used in mining topics from online reviews \citep{jo2011aspect,buschken2016sentence}.
Therefore, we select Amazon review data to perform the experiment. We first split each review sentence by ".", "!" and "?", and
then do the pre-processing for each sentence mentioned in Section \ref{SenLDA}. Finally, the data contains 17,216 reviews. We give the input file format of Sentence-LDA as follows:
\begin{Code}
love lotion--light clean smell
smell wonderful--feel great hand arm
good shoe
.
.
.
love comfy shoe--fit glove
\end{Code}
where the separator between two sentence in each review is "- -".

\pkg{TopicModel4J} offers a constructor method \code{SentenceLDA()}:
\begin{Code}
public SentenceLDA(String inputFile, String inputFileCode, int topicNumber,
                     double inputAlpha, double inputBeta, int inputIterations,
                       int inTopWords, String outputFileDir)
\end{Code}
The parameters (e.g., \code{inputFile} and \code{inputFileCode}) of this constructor method \code{SentenceLDA()} is the same as that of the constructor method \code{GibbsSamplingLDA()}.

The following code shows an example that extract topics from Amazon review data.
\begin{Code}
SentenceLDA sentenceLda = new SentenceLDA("data/amR_process", "gbk", 10, 0.1,
				0.01, 1000, 5, "data/senLDAoutput/");
sentenceLda.MCMCSampling();
\end{Code}
The output files (i.e., "SentenceLDA\_topic\_word10.txt" and "SentenceLDA\_doc\_topic\_10.txt") are saved to the output directory \code{"data/senLDAoutput/"}.
"SentenceLDA\_topic\_word10.txt" includes top 5 words and associated probabilities for each topic:
\begin{Code}
Topic:1
game :0.01275375756428569
vista :0.010205957681264213
xp :0.009926947895803648
windows :0.00936158596210724
user :0.008854962930613056

Topic:2
gift :0.012733120175753194
card :0.009475213711876724
love :0.00825349878792305
product :0.007371149120623172
great :0.007099656915300133

Topic:3
shoe :0.018754967742061632
comfortable :0.015284269721141634
wear :0.014974676939932808
foot :0.014909499512309897
size :0.013312652535548582
.
.
.
Topic:10
version :0.009524044352755055
program :0.008121748279908579
windows :0.007373857041057125
run :0.0068012528113114805
computer :0.006754509608883265
\end{Code}
"SentenceLDA\_doc\_topic\_10.txt" contains the topic distribution for each review, as shown below:
\begin{Code}
Topic1 Topic2 Topic3 ... Topic10
0.05 0.05 0.55 ... 0.05
0.05 0.55 0.05 ... 0.05
.
.
.
0.70 0.033 0.033 ... 0.033
\end{Code}

\subsubsection{Application of HDP}
In \pkg{TopicModel4J}, the constructor method \code{HDP()} is given as follows:
\begin{Code}
public HDP(String inputFile, String inputFileCode, int initTopicNumber,
	    double inputAlpha, double inputBeta, double inputGamma,
              int inputIterations, int inTopWords, String outputFileDir)
\end{Code}
where \code{inputFile} denotes the input data that its format is the same as that of \code{GibbsSamplingLDA()}.
The \code{initTopicNumber} is the initial topic number. The \code{inputAlpha}, \code{inputBeta} and \code{inputGamma} are
the input parameters $\alpha_{0}$, $\beta$ and $\gamma$ of HDP algorithm (See Algorithm 4).

In this case, we also use the citation data to illustrate the application of \code{HDP()}. The following code is to call the HDP algorithm for this data:
\begin{Code}
HDP hdp = new HDP("data/rawdataProcessAbstracts", "gbk", 3, 0.1, 0.01,
		  0.1, 1000, 5, "data/hdpoutput/");
hdp.MCMCSampling();
\end{Code}
We get outputs after running the code for \code{1000} iterations. In this experiment, we finally achieve the number of topics with $K=21$. The output directory \code{"data/hdpoutput/"} includes two files ("HDP\_topic\_word\_21.txt" and "HDP\_doc\_topic21.txt"). The contents of these two files are similar to the results of LDA.

\subsubsection{Application of DMM}
 The constructor method \code{DMM()} is given as follows:
\begin{Code}
public DMM(String inputFile, String inputFileCode, int clusterNumber,
           double inputAlpha, double inputBeta, int inputIterations,
                int inTopWords,String outputFileDir)
\end{Code}
The format of the \code{inputFile} is the same as that of \code{GibbsSamplingLDA()}. In the following, we perform the experiment using the Amazon review data. The code is given by:
\begin{Code}
DMM dmm = new DMM("data/amReviews", "gbk", 20, 0.1,
		0.01, 1000, 5, "data/dmmoutput/");
dmm.MCMCSampling();
\end{Code}
The \code{amReviews} contains 13,896 documents that are not consider the sentence segmentation. We set the cluster number at \code{20}. We will obtain the results after \code{1000} iterations. The output directory \code{"data/dmmoutput/"} includes three files ("DMM\_doc\_cluster20.txt", "DMM\_cluster\_word\_20.txt" and "DMM\_theta\_20.txt"). The file "DMM\_doc\_cluster20.txt" keeps the cluster assignment of each document, described as follows
\begin{Code}
8
8
5
13
.
.
.
8
\end{Code}

In this file, each line represents the cluster assignment of the document. For example,
the third document in \code{amReviews} is assigned to the cluster \code{5}.
The "DMM\_cluster\_word\_20.txt" gives the top 5 words for each topic:
\begin{Code}
Topic:1
windows :0.010503241790570985
program :0.009845820656453785
computer :0.009830167772308138
version :0.009485804321103892
work :0.00943884566866695

Topic:2
switch :0.010957253409633307
port :0.010174778755291485
hair :0.009913953870510878
love :0.00965312898573027
shampoo :0.009392304100949664

Topic:3
shoe :0.06688225340655453
comfortable :0.03773794324235105
fit :0.03644900569209144
size :0.03387113059157221
great :0.029359849165663563
.
.
.
Topic:20
gift :0.06384594555383939
card :0.05195676528401261
love :0.034568184448847454
great :0.031059043135726775
product :0.022574253692061848
\end{Code}

The "DMM\_theta\_20.txt" gives the weight of the mixture clusters estimated by \code{DMM()}:
\begin{Code}
0.031668705476002014
0.07916096999352377
0.0350507303734619
0.033539612866086206
0.0544074260631791
0.048650787939843135
0.1345686119306325
0.029294092250125928
0.02504857163416565
0.02497661365762395
0.05656616535943009
0.029294092250125928
0.0348348564438368
0.024400949845290353
0.012815715622076706
0.014326833129452399
0.051457149024969416
0.14824062747355543
0.028214722602000433
0.10355472404115995
\end{Code}

\subsubsection{Application of DPMM}
 In \pkg{TopicModel4J}, we provide the constructor method \code{DPMM()} as follows:
\begin{Code}
public DPMM(String inputFile, String inputFileCode, int initClusterNumber,
          double inputAlpha, double inputBeta, int inputIterations,
          int inTopWords,String outputFileDir)
\end{Code}
The \code{initClusterNumber} denotes the initial cluster number. An example implementation using DPMM is shown as follows:
\begin{Code}
DPMM dpmm = new DPMM("data/amReviews", "gbk", 3, 0.01,
				0.1, 1000, 5, "data/dpmmoutput/");
dpmm.MCMCSampling();
\end{Code}
We set the initial cluster number at 3. After 1000 iterations, we finally obtain the cluster number with 12. In the output directory \code{"data/dpmmoutput/"}, there are 3 files including "DPMM\_doc\_cluster12.txt", "DPMM\_cluster\_word\_12.txt" and "DPMM\_theta\_12.txt". The contents of these three files is similar to those of DMM.

\subsubsection{Application of PTM}
We provide the constructor method \code{PTM()} as follows:
\begin{Code}
public PseudoDTM(String inputFile, String inputFileCode, int pDocumentNumber,
         int topicNumber, double inputAlpha, double inputBeta, double inputLambada,
         int inputIterations, int inTopWords, String outputFileDir)
\end{Code}
where the \code{pDocumentNumber} is the number of pseudo document. The \code{inputAlpha}, \code{inputBeta} and \code{inputLambada} are
 the values of hyperparameters $\alpha$, $\beta$ and $\gamma$ in PTM. An example implementation using PTM is shown as follows:
\begin{Code}
PseudoDTM ptm = new PseudoDTM("data/amReviews", "gbk", 1000, 30, 0.1, 0.1,
				0.01, 1000, 5, "data/ptmoutput/");
ptm.MCMCSampling();
\end{Code}
We also apply \code{amReviews} to test PTM and set the \code{pDocumentNumber} at 1000. The topic number is fixed to 30. After 1000 iterations, the results including three files ("PseudoDTM\_topic\_word\_30.txt", "PseudoDTM\_pseudo\_topic30.txt" and "PseudoDTM\_doc\_topic30.txt") are saved to the output directory \code{"data/ptmoutput/"}. The "PseudoDTM\_topic\_word\_30.txt" gives the most frequent words for each topic:
\begin{Code}
Topic:1
powder :0.01467973081348722
brush :0.01223892046445134
clip :0.007357299766379581
exfoliate :0.00700861257366017
color :0.00700861257366017

Topic:2
map :0.015701847061823745
trips :0.014244599074647528
streets :0.013515975081059419
road :0.00950854311632482
gps :0.008415607125942658

Topic:3
shoe :0.07556278152996397
comfortable :0.030907744475191953
fit :0.02767832783033313
foot :0.027010172662431303
wear :0.02578522152127796
.
.
.
Topic:30
office :0.03880993933349954
user :0.016307717878398506
microsoft :0.01343101343101343
feature :0.013047452838028753
version :0.012855672541536417
\end{Code}
The "PseudoDTM\_doc\_topic30.txt" and "PseudoDTM\_doc\_topic30.txt" show the topic distribution for each document and pseudo document, respectively. The contents of these two files are similar to that of file "LDAGibbs\_doc\_topic30.txt" (See Application of LDA).

\subsubsection{Application of BTM}
We provide the constructor method \code{BTM()} as follows:
\begin{Code}
public BTM(String inputFile, String inputFileCode, int topicNumber,
        double inputAlpha, double inputBeta, int inputIterations, int inTopWords,
        int windowS, String outputFileDir)
\end{Code}
where the format of the \code{inputFile} is the same as that of \code{GibbsSamplingLDA()}. The \code{windowS} is the fixed-size sliding window that is used to generate biterms from the \code{inputFile}. Since BTM is suitable to fit short texts, we thus give an application example using \code{amReviews} data.
\begin{Code}
BTM btm = new BTM("data/amReviews", "gbk", 20, 0.1,
	          0.01, 1000, 30, 5, "data/btmoutput/");
btm.MCMCSampling();
\end{Code}
After 1000 iterations, the results including three files ("BTM\_topic\_word\_20.txt", "BTM\_topic\_theta\_20.txt" and "BTM\_doc\_topic\_20.txt") are saved to the output directory \code{"data/ptmoutput/"}.  The "BTM\_topic\_word\_20.txt" gives the most frequent words for each topic:
\begin{Code}
Topic:1
shoe :0.06670109601753213
wear :0.028623280599093498
foot :0.02479967286630265
comfortable :0.024707880958810154
size :0.02381844902758977

Topic:2
product :0.020062732717770964
software :0.01600907787414648
norton :0.013227409811881186
problem :0.013224197724049471
version :0.012305540604178947

Topic:3
soap :0.04160400014712544
scent :0.030190151053877717
smell :0.02769318031390717
skin :0.023470362150721687
bar :0.017099414748002716
.
.
.
Topic:20
file :0.019151633443044193
drive :0.018554723870893
dvd :0.013367495617128413
hard :0.01272855015961446
video :0.011702874556763115
\end{Code}
The "BTM\_topic\_theta\_20.txt" gives the result of the global topic distribution $\bm{\theta}$.
\begin{Code}
0.10303897002205605
0.1015355050050246
0.0747034832883077
0.03944514049306344
0.05327127624864597
0.028634417864087807
0.06805078109705963
0.021139279328661106
0.026122769925479294
0.03791622619839996
0.04357379637967712
0.030738224815003327
0.023991556060190804
0.017342116596844287
0.03456801482583558
0.050856205055923163
0.06774212703757358
0.08183123866202055
0.056731072915443145
0.038767798180702924
\end{Code}

The "BTM\_doc\_topic\_20.txt" presents the topic proportion for each document.
\begin{Code}
Topic1   Topic2   Topic3 ...  Topic20
4.572E-4 1.275E-4 0.281  ...  0.00733
0.0390   0.00469  1.713E-4 ... 1.890E-4
.
.
.
0.139   6.181E-4  3.262E-4 ... 2.697E-5
\end{Code}

\subsubsection{Application of ATM}
For ATM, we give an application example using citation data. From the citation data,
we first extract the author list and abstracts for each paper. For the abstracts, we then use the pre-processing method for the data cleaning. Finally, the input data for ATM contains 1,270 documents and its format is described as follows:
\begin{Code}
Christopher Kruegel, Engin Kirda, ...    number application download ...
Armando Barreto, Malek Adjouadi    paper present statistical result ...
Carlos T. Calafate, Juan-Carlos Cano, ...	multi-path route technique ...
.
.
.
Min Zhang, Shaoping Ma, ...	understand kind web page ...
\end{Code}
where any two authors in the author list of each paper are separated by a comma (\code{,}). The author list and abstracts are separated by Tab spaces (using "$\backslash$t" to create in Java).

We provide the constructor method \code{AuthorTM()} as follows:
\begin{Code}
public AuthorTM(String inputFile, String inputFileCode, String separator,
       int topicNumber, double inputAlpha, double inputBeta, int inputIterations,
        int inTopWords, String outputFileDir)
\end{Code}
where \code{separator} is the separator character of any two authors in the author list. An example implementation using ATM is shown as follows:
\begin{Code}
AuthorTM atm = new AuthorTM("data/citationAu_process", "gbk", ",", 30, 0.1,
				0.01, 1000, 5, "data/authorTMoutput/");
atm.MCMCSampling();
\end{Code}
Based on this implementation, we finally obtain the results including three files ("authorTM\_topic\_word30.txt", "authorTM\_author\_topic\_30.txt" and "authorTM\_topic\_author\_30.txt")
that are saved to the output directory \code{"data/authorTMoutput/"}.
The "authorTM\_topic\_word30.txt" presents the top 5 words for each topic:
\begin{Code}
Topic:1
graph :0.07544521097384886
rule :0.04775083465120365
network :0.041638972290344016
component :0.03686407982092243
edge :0.03361715294171575

Topic: 2
image :0.13050785410010649
generate :0.05111107941579187
multimedia :0.039938479385364364
training :0.03304474745169633
filter :0.022823006998326487
.
.
.
Topic:30
simulation :0.0762793382070027
robot :0.051134714431326486
sensor :0.050459690974261354
control :0.046578306096136834
logic :0.03932180393268666
\end{Code}
The "authorTM\_author\_topic\_30.txt" presents the distribution over topics for each author.
\begin{Code}
Christopher Kruegel 0.0172  2.132E-4 ... 0.0215
Engin Kirda 1.992E-4    1.992E-4 ...  1.992E-4
.
.
.
Franco Zambonelli  2.111E-4 0.0276 ...  0.554
\end{Code}
The "authorTM\_topic\_author\_30.txt" presents the top authors of each topic.
\begin{Code}
Topic:1
Gjergji Mino :0.6411764705882353
Schahram Dustdar :0.6065957446808511
Jian Ma :0.5407002188183807
Pietro Manzoni :0.42368972746331235
Francesco Marcelloni :0.40837789661319074

Topic:2
Yamin Li :0.6582
Jun He :0.4311340206185567
Song Wu :0.3729783037475345
Byeong-Soo Jeong :0.3082278481012658
Wanming Chu :0.26081330868761554
.
.
.
Topic:30
Min Huang :0.7492156862745099
Jie Lu :0.5713414634146342
Franco Zambonelli :0.5541226215644821
Laurence T. Yang :0.4674180327868852
Irith Pomeranz :0.42176470588235293
\end{Code}

\subsubsection{Application of Link LDA}
In this part, we use Link LDA to mine topics from the citation data. Based on the citation data, we first extract the citation relations and abstracts for each paper.
For the abstracts, we also use the pre-processing method for the data cleaning. Finally, the input data for Link LDA contains 18,299 documents and its format is described as follows:
\begin{Code}
457720--578743--6436976-- ...  paper present indoor navigation range ...
273266--1065537--11205936-- ...    globalisation education topic ...
210775--264701--272246-- ...  professional software development team base ...
.
.
.

\end{Code}
where any two citation indexes cited by each paper are separated by "- -". The citation indexes and abstracts are separated by Tab spaces. In \pkg{TopicModel4J}, the constructor method \code{LinkLDA()} is given as follows:
\begin{Code}
public LinkLDA(String inputFile, String inputFileCode, String separator,
         int topicNumber, double inputAlpha, double inputBeta,double inputGamma,
         int inputIterations, int inTopWords, String outputFileDir)
\end{Code}
where \code{separator} is the separator character of any two citation indexes. The \code{topicNumber}, \code{inputAlpha}, \code{inputBeta}, \code{inputGamma} and \code{inputIterations} are the
input parameters $\alpha$, $\beta$, $\gamma$, and $N_{iter}$ of Link LDA (See Algorithm 10). We give an example to show how to use Link LDA in our package.
\begin{Code}
LinkLDA linklda = new LinkLDA("data/citationLink_process", "gbk", "--", 50, 0.1,
				0.01,0.01, 1000, 5, "data/linkldaoutput/");
linklda.MCMCSampling();
\end{Code}
where we set the topic number at \code{50}. According to this code, we will obtain the results including three files ("LinkLDA\_topic\_word\_50.txt", "LinkLDA\_topic\_link\_50.txt" and "LinkLDA\_doc\_topic\_50.txt") that are saved to the output directory \code{"data/linkldaoutput/"}. The "LinkLDA\_topic\_word\_50.txt" gives the the most frequent words for each topic learned by Link LDA:
\begin{Code}
Topic:1
model :0.04919313426495349
distribution :0.031032042220064535
estimate :0.021740932811592357
parameter :0.019003608793232284
probability :0.018661443290937274

Topic:2
program :0.0361938918889249
language :0.023070572530579984
analysis :0.015965590712367767
code :0.01518543584605427
tool :0.014015203546584023
Topic:3
graph :0.08478070592614391
set :0.02273205662600254
vertex :0.021495890950958215
number :0.01871451818210849
edge :0.017066297282049395
.
.
.
Topic:50
datum :0.12070865610256701
rule :0.022095863862285934
database :0.02109702342522968
pattern :0.02046139769255752
mining :0.01888746730689312
\end{Code}
The "LinkLDA\_topic\_link\_50.txt" gives the the most influential node for each topic:
\begin{Code}
Topic:1
44875 :0.00726720143952733
891558 :0.0041754331180947285
129986 :0.002629548957378428
798508 :0.002629548957378428
891548 :0.002474960541306798

Topic:2
349116 :0.004260557418491948
297769 :0.003941095477590277
1415273 :0.0034086589094208257
809115 :0.0028762223412513747
473125 :0.0026632477139835946

Topic:3
408395 :0.002906343445558104
857281 :0.002583595700354539
219473 :0.0022608479551509734
836123 :0.0020994740825491906
1150707 :0.0017767263373456258
.
.
.
Topic:50
152933 :0.010751413698577275
481289 :0.010062308859545274
300119 :0.004273828211676468
818915 :0.004273828211676468
463902 :0.003171260469225267
\end{Code}
From the results in "LinkLDA\_topic\_word\_50.txt", we find that topic 50 is about pattern mining. Based on the citation indexes in "LinkLDA\_topic\_link\_50.txt", we can determine the most influential research papers from original citation data. In Table~\ref{tab:research paper}, we list the top 5 high-impact research papers related to topic 50.
\begin{table}[t!]
\centering
\begin{tabular}{lllp{7.4cm}}
\hline
Index    & Paper Title \\ \hline
152933     & Agrawal R, Imielinski T, Swami A. Mining association rules between sets of \\
           & items conference in large databases[C]//Proceedings of the 1993 ACM SIGMOD \\
           & international  on Management of data. 1993: 207-216. \\ \hline
481289     &  Agrawal R, Srikant R. Fast algorithms for mining association rules[C]//Proc. \\
           & 20th int. conf. very large data bases, VLDB. 1994, 1215: 487-499. \\ \hline
300119     & Han J, Pei J, Yin Y. Mining frequent patterns without candidate generation[J]. \\
           &  ACM sigmod record, 2000, 29(2): 1-12. \\ \hline
818915     & Han J, Pei J, Kamber M. Data mining: concepts and techniques[M].  \\
           & Elsevier, 2011. \\ \hline

463902     & Agrawal R, Srikant R. Mining sequential patterns[C]//Proceedings of the \\
           & eleventh international conference on data engineering. IEEE, 1995: 3-14. \\ \hline
\end{tabular}
\caption{\label{tab:research paper} Top research papers corresponding to topic 50.}
\end{table}

The "LinkLDA\_doc\_topic\_50.txt" shows the topic distribution for each document and its format is similar to that of LDA ("LDAGibbs\_doc\_topic30.txt").

\subsubsection{Application of Labeled LDA}
For Labeled LDA, we choose ProgrammableWeb data to perform the experiment. This ProgrammableWeb data contains 4,509 news and 81 label related to these news.
We give the input file format of Labeled LDA as follows:
\begin{Code}
Analysis,Collaboration  Microsoft confirm acquire repository ...
Security,Cloud  AWS handle credentials securely ...
Augmented Reality   announce general availability ...
.
.
.
Payments,Financial  effort establish partner choice global ...
\end{Code}
where any two labels are separated by a comma (\code{,}).
The label list and news content are separated by Tab spaces. In \pkg{TopicModel4J}, the constructor method \code{LabeledLDA()} is given as follows:
\begin{Code}
public LabeledLDA(String inputFile, String inputFileCode, String separator,
               double inputAlpha, double inputBeta, int inputIterations,
               int inTopWords, String outputFileDir)
\end{Code}
where \code{separator} is the separator character of any two labels. The \code{inputAlpha}, \code{inputBeta}, and \code{inputIterations} are the
input parameters  $\alpha$, and $\beta$ and $N_{iter}$  of Labeled LDA (See Algorithm 11).
Next, we give an example to show how to use this algorithm in \pkg{TopicModel4J}.
\begin{Code}
LabeledLDA llda = new LabeledLDA("data/programmableweb.txt", "gbk", ",", 0.1,
                        0.01, 1000, 5, "data/labeledLDAoutput/");
llda.MCMCSampling();
\end{Code}
After 1000 iterations, we get two output files, "LabeledLDA\_topic\_word\_81.txt" and "LabeledLDA\_doc\_topic81.txt", which are both
saved to the output directory \code{"data/labeledLDAoutput/"}. The "LabeledLDA\_topic\_word\_81.txt" presents the top 5 words for each topic:
\begin{Code}
Topic:1(Collaboration)
github :0.015267684231124786
feature :0.007339048698691607
project :0.006832965579600127
tool :0.006495576833539141
slack :0.006073840900962908

Topic:2(Storage)
storage :0.012778499109079106
dropbox :0.012433180932911585
file :0.008289362818901335
update :0.006217453761896211
aws :0.00569947649764493

Topic:3(COVID-19)
covid-19 :0.0334190259436744
coronavirus :0.01583638754319213
case :0.013870626728231384
health :0.01114040337411923
trace :0.007864135349184646
.
.
.
Topic:81(Augmented Reality)
reality :0.010398684912589058
game :0.008437039013213646
ar :0.008044709833338564
image :0.007848545243401022
augmented :0.007848545243401022
\end{Code}
The contents of "LabeledLDA\_doc\_topic81.txt" are similar to that of file "LDAGibbs\_doc\_topic30.txt" (See Application of LDA).

\subsubsection{Application of PLDA}
In \pkg{TopicModel4J}, the constructor method \code{PLDA()} is given as follows:
\begin{Code}
public PLDA(String inputFile, String inputFileCode, String separator,
         int label_topicNumber, double inputAlpha, double inputBeta,
         int inputIterations, int inTopWords, String outputFileDir)
\end{Code}

where \code{label_topicNumber} is the topic number of each label. And the \code{inputAlpha}, \code{inputBeta}, and \code{inputIterations}
are other input parameters $\alpha$, $\beta$ and $N_{iter}$ of PLDA (See Algorithm 12). In the following, we perform the experiment using the ProgrammableWeb data:
\begin{Code}
PLDA plda = new PLDA("data/programmableweb.txt", "gbk", ",", 2, 0.1, 0.01,
               1000, 20, "data/pLDAoutput/");
plda.MCMCSampling();
\end{Code}
We run the model with 2 global latent background topics and 2 latent topics per label. The results including two files ("PLDA\_topic\_word\_82.txt" and "PLDA\_doc\_topic82.txt") are
saved to the output directory \code{"data/pLDAoutput/"}.  The "PLDA\_topic\_word\_82.txt" shows the most frequent words for 164 (82*2) topics:
\begin{Code}
Topic:1	Related label:Collaboration
github :0.06933694647356305
feature :0.025834254744512314
slack :0.019036959161848138
collaboration :0.01813065308415958
project :0.017677500045315302

Topic:2	Related label:Collaboration
slack :0.022956225539933324
collaboration :0.020540416485480988
docusign :0.01933251195825482
evernote :0.018728559694641735
github :0.0163127506401894
.
.
.
Topic:163	Related label:global label
offer :0.0064778984884837
feature :0.005851266451286184
tool :0.0056916149768409564
enable :0.005116869668838139
system :0.0044982202053628845

Topic:164	Related label:global label
category :0.006814344167937558
offer :0.006224982169600439
tool :0.00542450542559032
feature :0.005016936680471616
client :0.004967090143298825
\end{Code}
The contents of "PLDA\_doc\_topic82.txt" are similar to that of file "LDAGibbs\_doc\_topic30.txt" (See Application of LDA).

\subsubsection{Application of Dual-Sparse Topic Model}
The Dual-Sparse Topic Model is a method for short documents modeling. In \pkg{TopicModel4J}, we create a constructor method \code{DualSparseLDA()} to implement this algorithm:
\begin{Code}
public DualSparseLDA(String inputFile, String inputFileCode, int topicNumber,
           double inputS, double inputT, double inputX, double inputY,
           double inputGamma,double inputGamma_bar,double inputPi,
           double inputPi_bar,int inputIterations, int inTopWords,
           String outputFileDir)
\end{Code}
where the \code{topicNumber}, \code{inputS}, \code{inputT}, \code{inputX}, \code{inputY}, \code{inputGamma}, \code{inputGamma_bar}, \code{inputPi}, \code{inputPi_bar} and  \code{inputIterations}  are the parameters of Dual-Sparse Topic Model (See Algorithm 13). We give an application example using Amazon review data of "Gift Cards" category. The data about this category contains 1,035 documents.
\begin{Code}
DualSparseLDA dualSLDA = new DualSparseLDA("data/amReviewsGift", "gbk", 10,
              1.0, 1.0, 1.0, 1.0, 0.1, 1E-12, 0.1, 1E-12, 1000, 5,
              "data/dstmoutput/");
dualSLDA.CVBInference();
\end{Code}

By this code, we obtain four output files "dualSLDA\_topic\_word\_10.txt", "dualSLDA\_doc\_topic\_10.txt", "dualSLDA\_sparseRatio\_TV10.txt", and "dualSLDA\_sparseRatio\_DT10.txt".
They are saved to the output directory \code{"data/linkldaoutput/"}. The "dualSLDA\_topic\_word\_10.txt" present the top 5 words for each topic:
\begin{Code}
Topic:1
love :0.05572016790679046
year :0.04240719351423932
daughter :0.031993038967226564
gift :0.03189584015804868
christmas :0.030028774929986162

Topic:2
gift :0.07582170588837109
card :0.06280373461792894
amazon :0.0408348675386099
balance :0.025295763275352393
box :0.025185954162112986

Topic:3
pizza :0.03393077466446514
restaurant :0.02224903253461759
food :0.02192767815153133
money :0.02086309815322604
save :0.020460047932054994
.
.
.
Topic:10
gift :0.120324455042531
card :0.08482478355137302
purchase :0.0643595870624188
family :0.023398387843003944
idea :0.02277221988826173
\end{Code}
The "dualSLDA\_doc\_topic\_10.txt" gives the topic distribution for each document and its format is similar to that of LDA ("LDAGibbs\_doc\_topic30.txt"). The "dualSLDA\_sparseRatio\_TV10.txt" shows the sparsity ratio of topic-word distributions and the average sparsity ratio of all topics.
\begin{Code}
0.9589682017129347
0.9408633633505831
0.944525877001114
0.9587098171129921
0.9516285406706164
0.9546314744349066
0.9608003344484427
0.9485143625733435
0.9562667091793843
0.9488602951277318
average saprse ratio of topic_word:0.9523768975612039
\end{Code}

 The "dualSLDA\_sparseRatio\_DT10.txt" shows the sparsity ratio of document-topic distributions and the average sparsity ratio of all documents.
 \begin{Code}
0.9999999999919068
0.5670308108642066
0.9999999999912975
.
.
.
0.9999999999827707
average saprse ratio of doc_topic:0.8804768437274924
\end{Code}

\subsection{Model Evaluation on Coherence Score}
For the probabilistic topic models, there are some metrics to evaluate model fit such as perplexity and coherence score \citep{lin2014dual}.
Perplexity is calculated by the inverse of the geometric mean per-word likelihood. The weakness of perplexity is that it disconnects
with human comprehension of topic quality \citep{chang2009reading}. Instead, \cite{mimno2011optimizing} propose an automatic metric, coherence score,
to evaluate semantic coherence of the learned topics.
Previous studies have indicated that this metric is correlated well with human judgments of topic quality \citep{mimno2011optimizing,newman2010automatic}.
Following \cite{mimno2011optimizing}, the coherence score for a topic $k$ is defined as:
\begin{equation} \label{eq:coherence}
C\left ( k; V^{\left ( k \right )} \right )=
\sum _{n=2}^{N}\sum _{l=1}^{n-1}log\frac{D\left (v_{n}^{\left ( k \right )},v_{l}^
{\left ( k \right )}  \right )+1 }{D\left ( v_{l}^{\left ( k \right )} \right )}
\end{equation}
where $V^{\left ( k \right )}=\left ( v_{1}^{\left ( k \right )},v_{2}^{\left ( k \right )},\cdots ,v_{N}^{\left ( k \right )} \right )$ represents the top $N$ words in topic $k$.
 $D\left ( v_{l}^{\left ( k \right )} \right )$ denotes the number of documents that occur $v_{l}^{\left ( k \right )}$.
$D\left (v_{n}^{\left ( k \right )},v_{l}^{\left ( k \right )}  \right )$ denotes the number of documents in which words $v_{n}^{\left ( k \right )}$
and $v_{l}^{\left ( k \right )}$ co-occurred.

In order to evaluate the overall quality, we use the average coherence score over all the topics:
\begin{equation} \label{eq:average coherence}
\overline{C}=\frac{1}{K}\sum _{k=1}^{K}C\left ( k; V^{\left ( k \right )} \right )
\end{equation}
The average coherence score has been widely used in evaluating the effectiveness of different topic models (e.g., LDA and BTM) or determining the number of topics for a
specific model (e.g., LDA) \citep{roberts2016model,xun2017collaboratively}. A higher coherence score indicates better model performance.
In \pkg{TopicModel4J}, we provide a \proglang{Java} method \code{average_coherence()} to implement the metric of coherence score:
\begin{Code}
public static double average_coherence(int[][] docs, double[][] phi,
          int topN_word)
\end{Code}
where \code{docs} denotes the encoding style (two dimensional array) of the input data. \code{phi} denotes topic-word distribution learned by a topic model.
\code{topN_word} denotes the number of top words in topics. We give an example to show the use of \code{average_coherence()}:
\begin{Code}
GibbsSamplingLDA lda = new GibbsSamplingLDA("data/rawdataProcessAbstracts",
          "gbk", 30, 0.1, 0.01, 500, 5, "data/ldagibbsoutput/");
lda.MCMCSampling();
double Ac5 = EstimationUtil.average_coherence(lda.docword, lda.estimatePhi(), 5);
double Ac10 = EstimationUtil.average_coherence(lda.docword, lda.estimatePhi(), 10);
double Ac20 = EstimationUtil.average_coherence(lda.docword, lda.estimatePhi(), 20);
System.out.println("average_coherence_5:\t" + Ac5);
System.out.println("average_coherence_10:\t" + Ac10);
System.out.println("average_coherence_20:\t" + Ac20);
\end{Code}
Running this code in a main method of \proglang{Java}, we can obtain the coherence score for the number of topic words $N=\left \{ 5,10,20 \right \}$.
The results is presented as follows.
\begin{Code}
average_coherence_5:	-17.377065821012504
average_coherence_10:	-88.70182574985039
average_coherence_20:	-392.8932343332532
\end{Code}

\section{Conclusion and Future Work}
Topic models play an prominent role for modeling unstructured text data.
This paper introduces a \proglang{Java} package,
named \pkg{TopicModel4J}, which implements multiple types of topic models.
For each topic model in \pkg{TopicModel4J}, we provide an easy-to-use interface to estimate the latent parameters of the model.
For the sake of convenient use for researchers, we give the detail description on the usage of \pkg{TopicModel4J} package, including data preparation and preprocessing, model estimation and model evaluation. From
usage examples, we find that each topic model in \pkg{TopicModel4J} can be easily implemented with a few line of \proglang{Java} code, and the results can be saved to
user specified file directory.

Package \pkg{TopicModel4J} is still undergoing active development. For the future work, we would like to add the visualization techniques
to this package for visualizing the results, e.g., generation of word clouds using topic-word distributions. In recent years,
there are some other variants of topic models developed by researchers. For example, \cite{blei2006correlated} propose a correlated topic model for
capturing rich topical correlations. \cite{blei2006dynamic} propose a dynamic topic model for tracking
the time evolution of topics in text data. \cite{wang2011collaborative} propose a collaborative topic modeling for recommendation tasks.
We plan to add these variants of topic models to \pkg{TopicModel4J}.

\section*{Acknowledgments}
This work is supported by the Major Program of the National Natural Science Foundation of China (91846201),
the Foundation for Innovative Research Groups of the
National Natural Science Foundation of China (71521001), the National Natural Science Foundation of China (71722010, 91746302, 71872060).


\bibliography{refs}

\begin{thebibliography}{44}
\newcommand{\enquote}[1]{``#1''}
\providecommand{\natexlab}[1]{#1}
\providecommand{\url}[1]{\texttt{#1}}
\providecommand{\urlprefix}{URL }
\expandafter\ifx\csname urlstyle\endcsname\relax
  \providecommand{\doi}[1]{doi:\discretionary{}{}{}#1}\else
  \providecommand{\doi}{doi:\discretionary{}{}{}\begingroup
  \urlstyle{rm}\Url}\fi
\providecommand{\eprint}[2][]{\url{#2}}

\bibitem[{Ansari \emph{et~al.}(2018)Ansari, Li, and
  Zhang}]{ansari2018probabilistic}
Ansari A, Li Y, Zhang JZ (2018).
\newblock \enquote{Probabilistic Topic Model for Hybrid Recommender Systems: A
  Stochastic Variational Bayesian Approach.}
\newblock \emph{Marketing Science}, \textbf{37}(6), 987--1008.

\bibitem[{Asuncion \emph{et~al.}(2009)Asuncion, Welling, Smyth, and
  Teh}]{asuncion2009smoothing}
Asuncion A, Welling M, Smyth P, Teh YW (2009).
\newblock \enquote{On smoothing and inference for topic models.}
\newblock In \emph{Proceedings of the twenty-fifth conference on uncertainty in
  artificial intelligence}, pp. 27--34. AUAI Press.

\bibitem[{Bi \emph{et~al.}(2014)Bi, Tian, Sismanis, Balmin, and
  Cho}]{bi2014scalable}
Bi B, Tian Y, Sismanis Y, Balmin A, Cho J (2014).
\newblock \enquote{Scalable topic-specific influence analysis on microblogs.}
\newblock In \emph{Proceedings of the 7th ACM international conference on Web
  search and data mining}, pp. 513--522. ACM.

\bibitem[{Blei and Lafferty(2006{\natexlab{a}})}]{blei2006correlated}
Blei D, Lafferty J (2006{\natexlab{a}}).
\newblock \enquote{Correlated topic models.}
\newblock \emph{Advances in neural information processing systems},
  \textbf{18}, 147.

\bibitem[{Blei and Lafferty(2006{\natexlab{b}})}]{blei2006dynamic}
Blei DM, Lafferty JD (2006{\natexlab{b}}).
\newblock \enquote{Dynamic topic models.}
\newblock In \emph{Proceedings of the 23rd international conference on Machine
  learning}, pp. 113--120.

\bibitem[{Blei \emph{et~al.}(2003)Blei, Ng, and Jordan}]{blei2003latent}
Blei DM, Ng AY, Jordan MI (2003).
\newblock \enquote{Latent dirichlet allocation.}
\newblock \emph{Journal of machine Learning research}, \textbf{3}(Jan),
  993--1022.

\bibitem[{B{\"u}schken and Allenby(2016)}]{buschken2016sentence}
B{\"u}schken J, Allenby GM (2016).
\newblock \enquote{Sentence-based text analysis for customer reviews.}
\newblock \emph{Marketing Science}, \textbf{35}(6), 953--975.

\bibitem[{Chang \emph{et~al.}(2009)Chang, Gerrish, Wang, Boyd-Graber, and
  Blei}]{chang2009reading}
Chang J, Gerrish S, Wang C, Boyd-Graber JL, Blei DM (2009).
\newblock \enquote{Reading tea leaves: How humans interpret topic models.}
\newblock In \emph{Advances in neural information processing systems}, pp.
  288--296.

\bibitem[{Cheng \emph{et~al.}(2014)Cheng, Yan, Lan, and Guo}]{cheng2014btm}
Cheng X, Yan X, Lan Y, Guo J (2014).
\newblock \enquote{Btm: Topic modeling over short texts.}
\newblock \emph{IEEE Transactions on Knowledge and Data Engineering},
  \textbf{26}(12), 2928--2941.

\bibitem[{Erosheva \emph{et~al.}(2004)Erosheva, Fienberg, and
  Lafferty}]{erosheva2004mixed}
Erosheva E, Fienberg S, Lafferty J (2004).
\newblock \enquote{Mixed-membership models of scientific publications.}
\newblock \emph{Proceedings of the National Academy of Sciences},
  \textbf{101}(suppl 1), 5220--5227.

\bibitem[{Gerlach \emph{et~al.}(2018)Gerlach, Peixoto, and
  Altmann}]{gerlach2018network}
Gerlach M, Peixoto TP, Altmann EG (2018).
\newblock \enquote{A network approach to topic models.}
\newblock \emph{Science advances}, \textbf{4}(7), eaaq1360.

\bibitem[{Gonz{\'a}lez-Blas \emph{et~al.}(2019)}]{gonzalez2019cistopic}
Gonz{\'a}lez-Blas CB, \emph{et~al.} (2019).
\newblock \enquote{cisTopic: cis-regulatory topic modeling on single-cell
  ATAC-seq data.}
\newblock \emph{Nature methods}, \textbf{16}(5), 397--400.

\bibitem[{Griffiths and Steyvers(2004)}]{griffiths2004finding}
Griffiths TL, Steyvers M (2004).
\newblock \enquote{Finding scientific topics.}
\newblock \emph{Proceedings of the National academy of Sciences},
  \textbf{101}(suppl 1), 5228--5235.

\bibitem[{Hannigan \emph{et~al.}(2019)}]{hannigan2019topic}
Hannigan T, \emph{et~al.} (2019).
\newblock \enquote{Topic modeling in management research: Rendering new theory
  from textual data.}
\newblock \emph{Academy of Management Annals}, (ja).

\bibitem[{Hornik and Gr{\"u}n(2011)}]{hornik2011topicmodels}
Hornik K, Gr{\"u}n B (2011).
\newblock \enquote{topicmodels: An R package for fitting topic models.}
\newblock \emph{Journal of Statistical Software}, \textbf{40}(13), 1--30.
\newblock \urlprefix\url{https://cran.r-project.org/web/packages/topicmodels}.

\bibitem[{Jo and Oh(2011)}]{jo2011aspect}
Jo Y, Oh AH (2011).
\newblock \enquote{Aspect and sentiment unification model for online review
  analysis.}
\newblock In \emph{Proceedings of the fourth ACM international conference on
  Web search and data mining}, pp. 815--824. ACM.

\bibitem[{Li \emph{et~al.}(2016)}]{li2016topic}
Li C, \emph{et~al.} (2016).
\newblock \enquote{Topic modeling for short texts with auxiliary word
  embeddings.}
\newblock In \emph{Proceedings of the 39th International ACM SIGIR conference
  on Research and Development in Information Retrieval}, pp. 165--174. ACM.

\bibitem[{Liang \emph{et~al.}(2016)}]{liang2016dynamic}
Liang S, \emph{et~al.} (2016).
\newblock \enquote{Dynamic clustering of streaming short documents.}
\newblock In \emph{Proceedings of the 22nd ACM SIGKDD international conference
  on knowledge discovery and data mining}, pp. 995--1004. ACM.

\bibitem[{Lin \emph{et~al.}(2014)Lin, Tian, Mei, and Cheng}]{lin2014dual}
Lin T, Tian W, Mei Q, Cheng H (2014).
\newblock \enquote{The dual-sparse topic model: mining focused topics and
  focused terms in short text.}
\newblock In \emph{Proceedings of the 23rd international conference on World
  wide web}, pp. 539--550. ACM.

\bibitem[{Ling \emph{et~al.}(2014)Ling, Lyu, and King}]{ling2014ratings}
Ling G, Lyu MR, King I (2014).
\newblock \enquote{Ratings meet reviews, a combined approach to recommend.}
\newblock In \emph{Proceedings of the 8th ACM Conference on Recommender
  systems}, pp. 105--112. ACM.

\bibitem[{Manning \emph{et~al.}(2014)Manning, Surdeanu, Bauer, Finkel, Bethard,
  and McClosky}]{manning2014stanford}
Manning CD, Surdeanu M, Bauer J, Finkel JR, Bethard S, McClosky D (2014).
\newblock \enquote{The Stanford CoreNLP natural language processing toolkit.}
\newblock In \emph{Proceedings of 52nd annual meeting of the association for
  computational linguistics: system demonstrations}, pp. 55--60.

\bibitem[{Mimno \emph{et~al.}(2011)Mimno, Wallach, Talley, Leenders, and
  McCallum}]{mimno2011optimizing}
Mimno D, Wallach H, Talley E, Leenders M, McCallum A (2011).
\newblock \enquote{Optimizing semantic coherence in topic models.}
\newblock In \emph{Proceedings of the 2011 Conference on Empirical Methods in
  Natural Language Processing}, pp. 262--272.

\bibitem[{Newman \emph{et~al.}(2010)Newman, Lau, Grieser, and
  Baldwin}]{newman2010automatic}
Newman D, Lau JH, Grieser K, Baldwin T (2010).
\newblock \enquote{Automatic evaluation of topic coherence.}
\newblock In \emph{Human language technologies: The 2010 annual conference of
  the North American chapter of the association for computational linguistics},
  pp. 100--108.

\bibitem[{Nigam \emph{et~al.}(2000)}]{nigam2000text}
Nigam K, \emph{et~al.} (2000).
\newblock \enquote{Text classification from labeled and unlabeled documents
  using EM.}
\newblock \emph{Machine learning}, \textbf{39}(2-3), 103--134.

\bibitem[{Phan and Nguyen(2007)}]{phan2007gibbslda++}
Phan XH, Nguyen CT (2007).
\newblock \enquote{GibbsLDA++: AC/C++ implementation of latent Dirichlet
  allocation (LDA).}
\newblock \emph{Tech. rep.}
\newblock \urlprefix\url{http://gibbslda.sourceforge.net/}.

\bibitem[{Ramage \emph{et~al.}(2009)Ramage, Hall, Nallapati, and
  Manning}]{ramage2009labeled}
Ramage D, Hall D, Nallapati R, Manning CD (2009).
\newblock \enquote{Labeled LDA: A supervised topic model for credit attribution
  in multi-labeled corpora.}
\newblock In \emph{Proceedings of the 2009 Conference on Empirical Methods in
  Natural Language Processing: Volume 1-Volume 1}, pp. 248--256. Association
  for Computational Linguistics.

\bibitem[{Ramage \emph{et~al.}(2011)Ramage, Manning, and
  Dumais}]{ramage2011partially}
Ramage D, Manning CD, Dumais S (2011).
\newblock \enquote{Partially labeled topic models for interpretable text
  mining.}
\newblock In \emph{Proceedings of the 17th ACM SIGKDD international conference
  on Knowledge discovery and data mining}, pp. 457--465. ACM.

\bibitem[{Roberts \emph{et~al.}(2016)Roberts, Stewart, and
  Airoldi}]{roberts2016model}
Roberts ME, Stewart BM, Airoldi EM (2016).
\newblock \enquote{A model of text for experimentation in the social sciences.}
\newblock \emph{Journal of the American Statistical Association},
  \textbf{111}(515), 988--1003.

\bibitem[{Rosen-Zvi \emph{et~al.}(2004)Rosen-Zvi, Griffiths, Steyvers, and
  Smyth}]{rosen2004author}
Rosen-Zvi M, Griffiths T, Steyvers M, Smyth P (2004).
\newblock \enquote{The author-topic model for authors and documents.}
\newblock In \emph{Proceedings of the 20th conference on Uncertainty in
  artificial intelligence}, pp. 487--494. AUAI Press.

\bibitem[{Sethuraman(1994)}]{sethuraman1994constructive}
Sethuraman J (1994).
\newblock \enquote{A constructive definition of Dirichlet priors.}
\newblock \emph{Statistica sinica}, pp. 639--650.

\bibitem[{Su \emph{et~al.}(2018)Su, Wang, Zhang, Chang, and
  Zia}]{su2018identifying}
Su S, Wang Y, Zhang Z, Chang C, Zia MA (2018).
\newblock \enquote{Identifying and tracking topic-level influencers in the
  microblog streams.}
\newblock \emph{Machine Learning}, \textbf{107}(3), 551--578.

\bibitem[{Teh \emph{et~al.}(2005)Teh, Jordan, Beal, and Blei}]{teh2005sharing}
Teh YW, Jordan MI, Beal MJ, Blei DM (2005).
\newblock \enquote{Sharing clusters among related groups: Hierarchical
  Dirichlet processes.}
\newblock In \emph{Advances in neural information processing systems}, pp.
  1385--1392.

\bibitem[{Teh \emph{et~al.}(2007)Teh, Newman, and Welling}]{teh2007collapsed}
Teh YW, Newman D, Welling M (2007).
\newblock \enquote{A collapsed variational Bayesian inference algorithm for
  latent Dirichlet allocation.}
\newblock In \emph{Advances in neural information processing systems}, pp.
  1353--1360.

\bibitem[{Wang and Blei(2011)}]{wang2011collaborative}
Wang C, Blei DM (2011).
\newblock \enquote{Collaborative topic modeling for recommending scientific
  articles.}
\newblock In \emph{Proceedings of the 17th ACM SIGKDD international conference
  on Knowledge discovery and data mining}, pp. 448--456.

\bibitem[{Wang \emph{et~al.}(2015)Wang, Lee, Chin, Chen, and
  Hsieh}]{wang2015hierarchical}
Wang JC, Lee YS, Chin YH, Chen YR, Hsieh WC (2015).
\newblock \enquote{Hierarchical Dirichlet process mixture model for music
  emotion recognition.}
\newblock \emph{IEEE Transactions on Affective Computing}, \textbf{6}(3),
  261--271.

\bibitem[{Xun \emph{et~al.}(2017)Xun, Li, Gao, and
  Zhang}]{xun2017collaboratively}
Xun G, Li Y, Gao J, Zhang A (2017).
\newblock \enquote{Collaboratively improving topic discovery and word
  embeddings by coordinating global and local contexts.}
\newblock In \emph{Proceedings of the 23rd ACM SIGKDD International Conference
  on Knowledge Discovery and Data Mining}, pp. 535--543.

\bibitem[{Yan \emph{et~al.}(2013)Yan, Guo, Lan, and Cheng}]{yan2013biterm}
Yan X, Guo J, Lan Y, Cheng X (2013).
\newblock \enquote{A biterm topic model for short texts.}
\newblock In \emph{Proceedings of the 22nd international conference on World
  Wide Web}, pp. 1445--1456. ACM.

\bibitem[{Yang \emph{et~al.}(2009)Yang, Jin, Chi, and Zhu}]{yang2009combining}
Yang T, Jin R, Chi Y, Zhu S (2009).
\newblock \enquote{Combining link and content for community detection: a
  discriminative approach.}
\newblock In \emph{Proceedings of the 15th ACM SIGKDD international conference
  on Knowledge discovery and data mining}, pp. 927--936. ACM.

\bibitem[{Yin and Wang(2014)}]{yin2014dirichlet}
Yin J, Wang J (2014).
\newblock \enquote{A dirichlet multinomial mixture model-based approach for
  short text clustering.}
\newblock In \emph{Proceedings of the 20th ACM SIGKDD international conference
  on Knowledge discovery and data mining}, pp. 233--242. ACM.

\bibitem[{Yin and Wang(2016)}]{yin2016model}
Yin J, Wang J (2016).
\newblock \enquote{A model-based approach for text clustering with outlier
  detection.}
\newblock In \emph{2016 IEEE 32nd International Conference on Data Engineering
  (ICDE)}, pp. 625--636. IEEE.

\bibitem[{Yu \emph{et~al.}(2010)}]{yu2010document}
Yu G, \emph{et~al.} (2010).
\newblock \enquote{Document clustering via dirichlet process mixture model with
  feature selection.}
\newblock In \emph{Proceedings of the 16th ACM SIGKDD international conference
  on Knowledge discovery and data mining}, pp. 763--772. ACM.

\bibitem[{Yuan \emph{et~al.}(2014)Yuan, Cong, and Lin}]{yuan2014generative}
Yuan Q, Cong G, Lin CY (2014).
\newblock \enquote{COM: a generative model for group recommendation.}
\newblock In \emph{Proceedings of the 20th ACM SIGKDD international conference
  on Knowledge discovery and data mining}, pp. 163--172. ACM.

\bibitem[{Zhang \emph{et~al.}(2005)Zhang, Ghahramani, and
  Yang}]{zhang2005probabilistic}
Zhang J, Ghahramani Z, Yang Y (2005).
\newblock \enquote{A probabilistic model for online document clustering with
  application to novelty detection.}
\newblock In \emph{Advances in neural information processing systems}, pp.
  1617--1624.

\bibitem[{Zuo \emph{et~al.}(2016)}]{zuo2016topic}
Zuo Y, \emph{et~al.} (2016).
\newblock \enquote{Topic modeling of short texts: A pseudo-document view.}
\newblock In \emph{Proceedings of the 22nd ACM SIGKDD international conference
  on knowledge discovery and data mining}, pp. 2105--2114. ACM.

\end{thebibliography}


\newpage


\end{document}